\documentclass[letterpaper]{article} % DO NOT CHANGE THIS
\usepackage{aaai24}  % DO NOT CHANGE THIS

\usepackage{times}  % DO NOT CHANGE THIS
\usepackage{helvet}  % DO NOT CHANGE THIS
\usepackage{courier}  % DO NOT CHANGE THIS
\usepackage[hyphens]{url}  % DO NOT CHANGE THIS
\usepackage{graphicx} % DO NOT CHANGE THIS
\usepackage{natbib}  % DO NOT CHANGE THIS AND DO NOT ADD ANY OPTIONS TO IT
\usepackage{caption} % DO NOT CHANGE THIS AND DO NOT ADD ANY OPTIONS TO IT
\usepackage{amssymb}% http://ctan.org/pkg/amssymb
\usepackage{pifont}% http://ctan.org/pkg/pifont
\usepackage{algorithm}
\usepackage{algorithmic}

\usepackage{newfloat}
\usepackage{listings}
\usepackage{subcaption,siunitx,booktabs}
\usepackage{sidecap, caption}
\DeclareCaptionStyle{ruled}{labelfont=normalfont,labelsep=colon,strut=off} % DO NOT CHANGE THIS
\floatstyle{ruled}
\newfloat{listing}{tb}{lst}{}
\floatname{listing}{Listing}
\usepackage{booktabs,makecell}
\usepackage{amssymb}
\usepackage{color}
\usepackage{sidecap,wasysym,array,tabularx,caption}
\definecolor{atomictangerine}{rgb}{0.8, 0.2, 0.1}
\definecolor{turq}{rgb}{0.0, 0.5, 0.5}
\definecolor{darkturq}{rgb}{0.0, 0.4, 0.4}
\definecolor{bright}{rgb}{0.8, 0.1, 0}
\definecolor{darkgray}{gray}{0.3}
\definecolor{mahogany}{rgb}{0.6, 0.05, 0.05}
\definecolor{myblue}{rgb}{0.3,0.05,0.9}

\definecolor{olive}{rgb}{0.537, 0.627, 0.318}
\definecolor{green}{rgb}{0.22, 0.463, 0.114}
\definecolor{grey}{rgb}{0.4, 0.4, 0.4}
\definecolor{blue}{rgb}{0.435, 0.659, 0.863}
\definecolor{pink}{rgb}{0.761, 0.482, 0.627}
\definecolor{darkpink}{rgb}{0.561, 0.282, 0.427}

\usepackage{soul}

\usepackage[capitalize]{cleveref}
\newcommand{\ourmethod}{{SeedSelect}}
\newcommand\NSO{\ourmethod{}}
\newcolumntype{H}{>{\setbox0=\hbox\bgroup}c<{\egroup}@{}}

\nocopyright %-- Your paper will not be published if you use this command
\title{Generating images of rare concepts using pre-trained diffusion models
}
\author{
    Dvir Samuel\textsuperscript{\rm 1,2}\thanks{Correspondence to: Dvir Samuel $<$dvirsamuel@gmail.com$>$}, Rami Ben-Ari\textsuperscript{\rm 2}, Simon Raviv\textsuperscript{\rm 1}, Nir Darshan\textsuperscript{\rm 2}, Gal Chechik\textsuperscript{\rm 1,3}
}
\affiliations{
    \textsuperscript{\rm 1}Bar-Ilan University, Ramat-Gan, Israel\\
    \textsuperscript{\rm 2}OriginAI, Tel-Aviv, Israel\\
    \textsuperscript{\rm 3}NVIDIA Research, Tel-Aviv, Israel\\
}

\begin{document}
\pagestyle{plain} % Remove before submission
\maketitle

\begin{abstract}
Text-to-image diffusion models can synthesize high quality images, but they have various limitations. Here we highlight a common failure mode of these models, namely, generating uncommon concepts and structured concepts like hand palms. We show that their limitation is partly due to the long-tail nature of their training data: web-crawled data sets are strongly unbalanced, causing models to under-represent concepts from the tail of the distribution. We characterize the effect of unbalanced training data on text-to-image models and offer a remedy. We show that rare concepts can be correctly generated by carefully selecting suitable generation seeds in the noise space, using a small reference set of images, a technique that we call \ourmethod{}. \ourmethod{} does not require retraining or finetuning the diffusion model.
We assess the faithfulness, quality and diversity of \ourmethod{} in creating rare objects and generating complex formations like hand images, and find it consistently achieves superior performance. We further show the advantage of \ourmethod{} in semantic data augmentation. Generating semantically appropriate images can successfully improve  performance in few-shot recognition benchmarks, for classes from the head and from the tail of the training data of diffusion models.

\end{abstract}

\section{Introduction}

\begin{figure}[t!]
\centering
    \includegraphics[width=1\columnwidth]{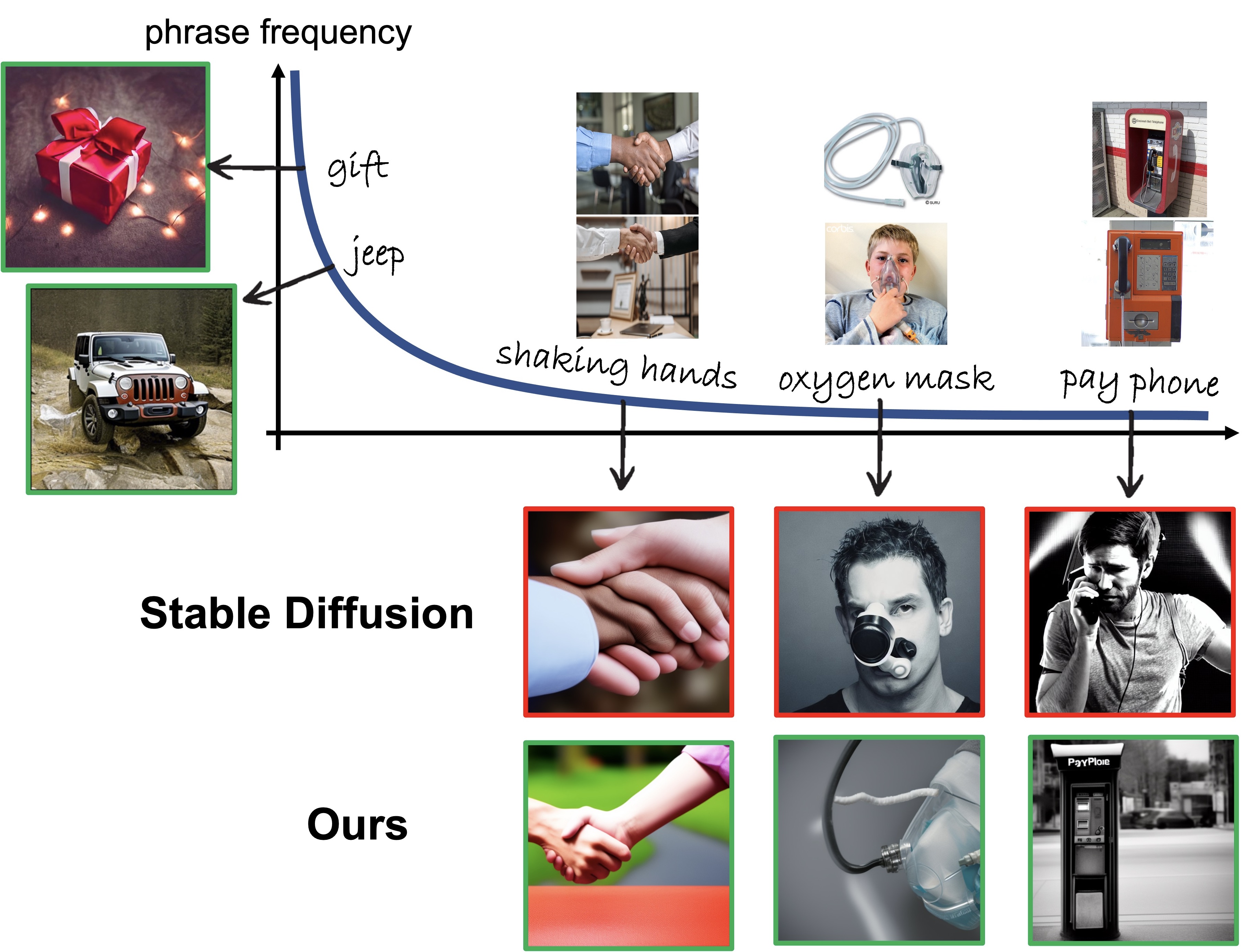}
    \caption{Generating rare concepts. Current diffusion models fail when conditioned on phrases or classes that are in the tail of their training distribution, like \textit{pay phone}, or structurally complex classes like \textit{shaking hands}. \ourmethod{} fixes that using just a handful of additional references images, without any fine-tuning.
    }
    \label{figure_1}
\end{figure}

Diffusion models achieve unprecedented success in text-to-image generation. They map a noise vector sampled from a high-dimensional Gaussian, conditioned on a text prompt, to a corresponding image \cite{StableDiffusion,saharia2022photorealistic,ramesh2022hierarchical,balaji2022ediffi}. While successful, several failure modes of current models have been identified. Common failures range from omitting objects listed in the prompt or confusing their attributes \cite{chefer2023attend, Rassin2023LinguisticBI}, through ignoring spatial relations \cite{lian2023llm} to generating deformed hands as illustrated in Figure \ref{figure_1}.

One failure mode received less attention so far: some concepts and object classes consistently fail to be drawn correctly. Figure 1 illustrates these failures for two concepts: ``pay phone" and ``oxygen mask" in images generated with  StableDiffusion \cite{StableDiffusion}. These failures occur mostly with concepts that appear less frequently, but it is still not well understood what causes these failures, and if at all they can be corrected.

Here we study a major failure mode of text-to-image diffusion models: generation of concepts that are under-represented in the training data. We first quantify this effect in a public model (Stable Diffusion) trained with public data \citep[Laion2B]{Laion5B_dataset}, and find  that 25\% of ImageNet concepts are poorly generated (Figure 2). The failing concepts are those that have fewer than 10K samples in the training data of  the diffusion model.
This observation is somewhat puzzling. Intuitively, 10k samples should be sufficient for learning the appearance of a concept, even a complex one.

Why do diffusion models fail to generate images from concepts with several thousand image samples? One possible answer raises from failures of deep models trained for \textit{long-tail} recognition \cite{zhang2021deep}. There, common concepts dominate the learned representation, washing out the representation of rare concepts. If this type of  "catastrophic forgetting" is the cause for the above failures in generative models, little can be done to improve the generation of rare concepts.

This paper explores a different answer.
Our insight is that diffusion models may be sensitive to the initial random noise used as input in conjunction with their text prompts. When a diffusion model is trained for frequent concepts ("A dog"), its training covers a large fraction of the random input space. The model then learns to map any noise sample correctly to viable images. In contrast, for rare concepts, only a small fraction of that input space is observed during training. As a result, at generation time for a given prompt, the model may view many random inputs as out-of-distribution.

Based on this view, we show that, indeed, diffusion models can generate images from rare concepts, as long as the initial noisy image (the seed) is carefully chosen. To achieve this, we use a small set of reference images from the class. We identify areas in the noise space (seeds) that would be ``in-distribution" for the diffusion model for a given prompt. More concretely, we do a gradient-based search in input space for regions that generate images that are similar, visually and semantically,  to our few-shot reference  set.
We call our approach \textbf{\ourmethod{}}.

We evaluate the quality of images generated with \ourmethod{} in several ways.
First, we evaluate the faithfulness of generation, namely, if generated images depict the correct class. This is done (a) using a classifier that was pre-trained to recognize each concept and (b) using human raters. \ourmethod{} consistently achieves better faithfulness than all competing approaches, for concepts that are in the tail of the Laion2B distribution, across three datasets (Imagnet, iNaturalist and CUB). \ourmethod{} also achieves better image quality as measured using FID.
Then, we test the benefit of using generated images for semantic augmentation in an object recognition task.  \ourmethod{}  achieves state-of-the-art results for few-shot image recognition tasks on ImageNet, CUB, and iNaturalist. It generates valuable, diverse, and superior augmentations compared to previous methods.
Finally, \ourmethod{} can be used to improve generation of challenging concepts, such as hand palms, where current diffusion models struggle.

Our paper makes the following contributions:
(1) We characterize  the failure of text-to-image diffusion models to generate images of rare concepts.
(2) We introduce the learning setup of rare-concept generation using a reference set and a pre-trained text-to-image diffusion model.
(3) We describe \textit{\ourmethod{}}, a novel method to improve generation of uncommon and ill-formed concepts in diffusion models. It operates as per-class test-time optimization by finding a generation seed from just a few reference samples.
(4) We propose an efficient bootstrapping technique to accelerate image generation with \ourmethod{}.

\section{Related Work}

\noindent\textbf{Text-guided generation:}  Diffusion models provide unprecedented quality for text-to-image generation \cite{ramesh2022hierarchical,saharia2022photorealistic,balaji2022ediffi} but still struggle with rare fine-grained objects and compositions \cite{chefer2023attend,compositionalDM_ECCV22}. Techniques like pre-trained image classifiers \cite{dhariwal2021diffusion} and text-driven gradients \cite{ho2022classifier, nichol2021glide,saharia2022photorealistic} have been proposed for better aligning generated images with the given text prompt, but require pre-trained classifier which may not be available or extensive prompt engineering \cite{liu2022design,marcus2022very,wang2022diffusiondb}. Other approaches \cite{avrahami2022spatext,feng2022training,chefer2023attend}  generate more accurate images, and focus on aligning better the generated images to the prompt, not addressing the generation of rare objects. Our approach also improves alignment with the prompt, in the sense of forcing the model to generate a well-formed or correct image, particularly when the concept is rare.

\paragraph{Semantic  augmentations for image recognition with pre-trained text-to-image models:}
Recently, \cite{He2022IsSD,FTSD2023} showed that data augmentations obtained from images generated by pre-trained text-to-image models improve zero-shot and few-shot image classification. \cite{He2022IsSD} achieves SOTA results by fine-tuning a CLIP classifier with real and synthetic images. Two strategies were introduced for generating images resembling few-shot reference images: (1) Real Guidance (RG) guides image generation using few-shot real samples, where these samples (with added noise) replace initial random noise to steer the diffusion process. (2)
Real Filtering (RF) uses few-shot real sample features to filter similar synthetic images.
However effective, we show that these strategies compromise image diversity and naturalness and aren't suitable for generating rare concepts. \cite{FTSD2023} showed that large-scale text-to-image diffusion models, when fine-tuned, can produce class conditional models that enable classifiers trained on such generated data to excel in classification benchmarks. Despite their effectiveness, this approach demands substantial fine-tuning data, which might be lacking, especially for generating rare concepts. Our approach, on the other hand, demonstrates how to generate such rare concepts without finetuning the diffusion model.

\paragraph{Image generation personalized to an instance:}
Recently, \cite{gal2022image, ruiz2022dreambooth, Tewel2023KeyLockedRO} described how few reference samples can be used to train a model to generate images of a unique instance object. In principle, these methods can also be used for generating rare concepts and for few-shot semantic data-augmentation.
However, they require long-time training for a single concept, and importantly, they do not learn a "class concept" but rather an "instance-specific concept" (or style), as in "this \textit{specific} cat" and not "this \textit{type} of cat". Accelerated versions like \cite{gal2023designing} are limited to specific classes. Overall, these methods require substantial computational resources. Thus, when evaluating them as data augmentation methods, we only compare our approach with Textual Inversion~\cite{gal2022image} on the CUB~\cite{CUB} dataset.

\section{Motivating Analysis}
\label{sec:motivation}
We start by quantifying the relation between two quantities: the faithfulness of images generated for a given class by a common text-to-image model, and the number of samples from that class in the training set.

Foundation diffusion models are trained on massive datasets, collected "in the wild" from the web \cite{Laion5B_dataset}. The distribution of concepts in web images is highly unbalanced, with some concepts appearing orders-of-magnitude more frequently than others. As a result, trained diffusion models are well-tuned to ``head" concepts, but when asked to generate images from ``tail" classes, the results are poor.
Figure \ref{figure_1} illustrates how this imbalance is manifested in the LAION2B dataset~\cite{Laion5B_dataset}. We parsed all image captions and extracted all noun phrases in each caption (more details and a similar analysis of LAION400M are given in the supplementary. Data will be publicly released).

\begin{figure}[t!]
    \centering
    \includegraphics[width=0.8\columnwidth]{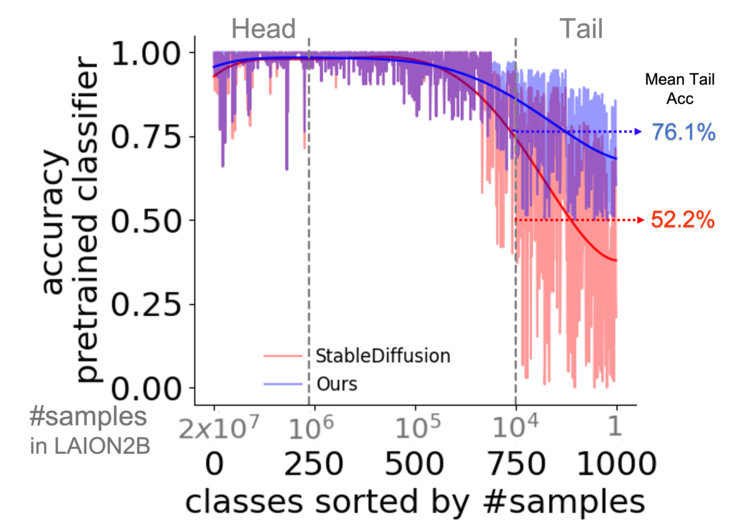}
    \caption{
    Per-class accuracy of a pre-trained classifier for images generated using stable diffusion. Shown are the 1000 classes of ImangeNet1k ordered by their number of occurrences in the LAION2B dataset.}
    \label{fig:laion}
\end{figure}

We quantified the relation between faithfulness and training imbalance with the following experiment. For every class in ImageNet~\cite{ImageNet}, we used Stable Diffusion \citep[SD]{StableDiffusion} to generate 100 images using the class label as the prompt.  We then used a SoTA pre-trained classifier provided by \citet{tu2022maxvit} to test if the generated images are from the correct class (see supplementary for details). That classifier was trained on balanced ImageNet data and has no preference for classes that appear at the head of the Laion distribution.
Figure \ref{fig:laion} depicts the resulting per-class accuracy for ImageNet classes sorted by their prevalence in the LAION2B dataset. Images generated for categories at the head of LAION2B distribution yield high accuracy, but accuracy drops significantly at the tail, particularly for classes in the last quartile (last 25\% of classes). For those rare classes, about 50\% of synthetic images generated by Stable Diffusion are correctly identified, indicating corrupted or incorrect concepts. In the figure, we also report the mean accuracy of classes from the last quartile (mean tail acc).
Note that concepts from many tail classes were observed thousands of times in the Laion training data. This behavior strongly limits the usability of diffusion models to generate rare concepts.

Since the diffusion model was trained with thousands of samples from rare classes, a natural question arises: \textbf{Are these classes encoded in the model? and if so can they be revived and generated?} or were they washed out by the overwhelmingly many more samples from head classes?

\textbf{Our working hypothesis:}
Deep diffusion models are trained given two inputs: a text prompt, and a noisy image, which in the extreme case is a noise tensor sampled from a high-dimensional Gaussian distribution.
We propose that when trained with common (head) concepts, the model learns to map large parts of that Gaussian distribution into images of correct concepts. However, for rare (tail) concepts, the model can generate correct concepts only for limited areas of that distribution.  If that is true, then if we can locate these areas of the distribution, we could still generate images of rare concepts. In this paper, we propose to discover these areas by optimizing over the seed in the noise-space, such that it improves semantic and appearance agreement with a small set of reference images of target rare concepts. Figure \ref{fig:laion} shows that images generated by our approach achieve better faithfulness. In the subsequent sections, we will elaborate on the details of our method.

\section{Notations and definitions}
\label{sec:notation}

We start with defining the problem of rare-concept generation with a reference set. Given a pre-trained text-to-image model (like StableDiffusion), a rare concept $y$ to be generated, and $k$ reference images $I^{1}, I^{2}, ... I^{k}$ of the concept $y$, the objective is to generate new and semantically-correct images of $y$.

While our approach can be directly applied to all diffusion models, in this work we use the open-sourced model of Stable Diffusion (SD) ~\cite{StableDiffusion}. In the context of SD, a denoising diffusion probabilistic model (DDPM) is applied to the latent space of a variational auto-encoder. The process involves training an encoder $\mathcal{E}$ to map images to spatial latent codes $z$, and a decoder $\mathcal{D}$ to reconstruct images from these codes. The DDPM, informed by conditioning vectors (often derived from pre-trained CLIP text encoders), operates on the latent space and uses a network $\varepsilon_\theta$ to effectively remove noise $\varepsilon$ from the latent code $z$ using UNet architecture with self-attention and cross-attention layers. During inference, a latent $z_T$ is sampled from a standard normal distribution and iteratively denoised with DDPM to yield a latent $z_0$, which is then decoded by $\mathcal{D}$ to generate the final image $I^{G}$. More details in Supp.

\section{Our Approach: Seed Select}

\begin{figure}[t!]
    \centering     \includegraphics[width=1\columnwidth]{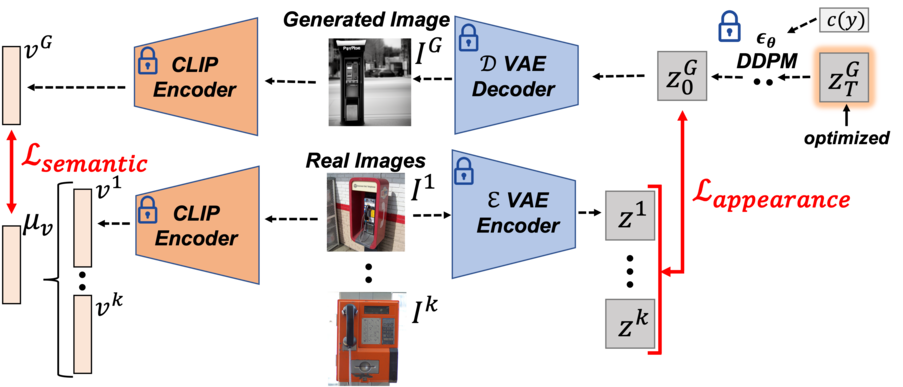}
    \caption{An overview of \ourmethod{}. An initial noise $z_T^G$ is used to generate an image $I^G$. It is then tuned to minimize a semantic loss (using clip image encoder) and an appearance loss (using the diffusion VAE) based on its match to reference samples $I^1,\ldots, I^k$.
    }
    \label{fig:method}
\end{figure}

We now describe how we use {\it few reference images}, $I^1,\ldots, I^k$,  to improve generation of images for a given prompt $y$. Typically, $k$ can be set to 3-5 samples. Our goal is to find an initial noise tensor $z^G_T$ that generates images that are consistent with the reference set as illustrated in Figure \ref{fig:method}.
We measure this consistency in two ways:

\textbf{(1) Semantic consistency.} Measures the semantic similarity between the generated image $I^G$ obtained from a seed $z_{T}^{G}$ and the reference images $I^1,\ldots,I^k$. Specifically,
we use a pre-trained CLIP image encoder to encode the reference images into $v^1,\ldots v^k$, and compute their centroid (mean vector): $\mu_{v} = mean(v^1,\ldots,v^k)$. Similarly, we encode the generated image and obtain $v^{G}$.  The semantic loss is then:
\begin{equation}
    \mathcal{L}_{Semantic} = dist_{v}(\mu_{v}, v^{G}),
\end{equation}
where $dist_{v}$ is the euclidean distance between the
centroid $\mu_{v}$ and a feature vector $v^{G}$. This loss makes sure that the semantic concept in the generated image corresponds to the concept presented in the reference images.

\textbf{(2) Natural appearance consistency.}
Measures the similarity between the spatial latent $z_{0}^{G}$ obtained from $z_{T}^{G}$ during the denoising process and the encoded reference images. More specifically, we encode the reference images $I^1,\ldots,I^k$ using $\mathcal{E}$, the VAE encoder, to obtain $z^1,\ldots,z^k$.
Then, we define the appearance loss to be:
\begin{equation}
    \mathcal{L}_{Appearance} = \frac{1}{k} \sum_{i=1}^{k}{ dist_{z}(z^{i}, z_{0}^{G} )},
\end{equation}
where $dist_{z}$ is pixel-wise mean-squared error loss. Note that this loss mirrors the loss used during training of the diffusion model, namely the MSE between the latent tensor of the generated image and the latent representation of the provided real images. This loss mechanism ensures that the generated images maintain a natural appearance consistent with the provided images.

The overall loss is then:
\begin{equation}
    \mathcal{L}_{Total} = \lambda \mathcal{L}_{Semantic} + (1-\lambda) \mathcal{L}_{Appearance} \quad.
\end{equation}
Here, $\lambda$ is a hyperparameter to control the tradeoff between semantic and appearance.

We only optimize $z_{T}^{G}$, the initial generation point, by backpropagating the loss through the denoising model, maximizing appearance and semantic consistency.
\newline

\begin{figure*}[t!]
    \centering     \includegraphics[width=1.7\columnwidth]{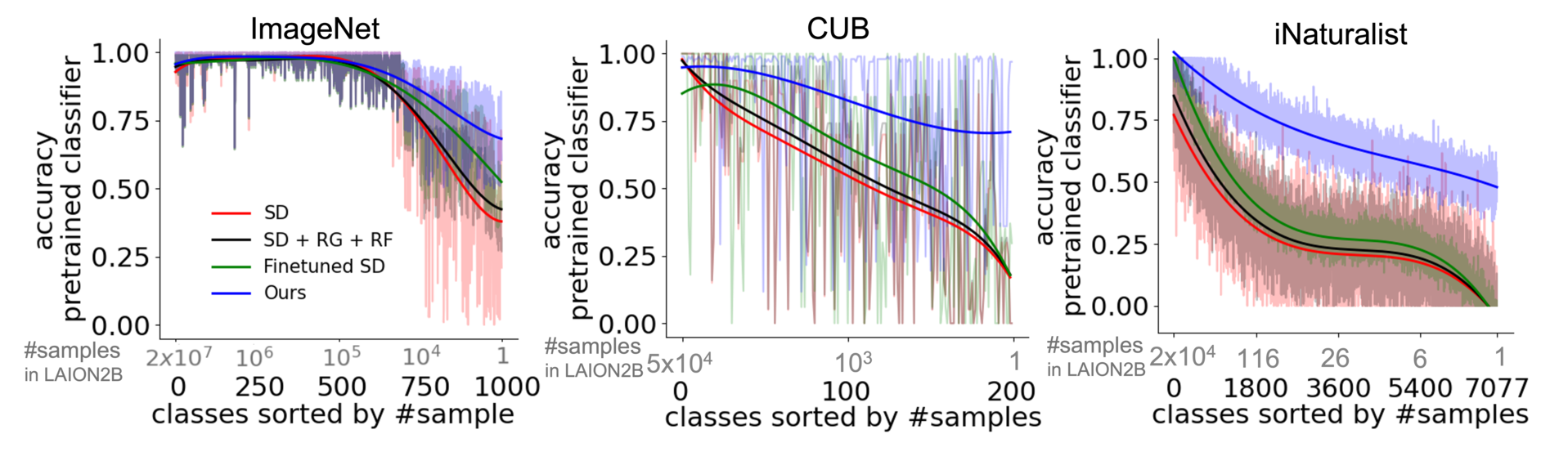}
    \caption{Per-class accuracy of pre-trained object recognition  given images generated using various approaches. Classes are ordered by their number of occurrences in LAION2B. \ourmethod{} achieves the highest accuracy for all classes across all benchmarks, outperforming previous methods. Corresponding tables can be found in Supp. Solid lines: Polynomial fits.
    }
    \label{fig:rare_eval}
\end{figure*}

\noindent\textbf{Implementation details:} See the supplemental. \newline

\noindent\textbf{Stopping criteria:} We stop optimizing $z_{T}^G$ when $\mathcal{L}_{Total}$ plateaus or its value increases for more than 3 iterations.
\newline

\noindent\textbf{Inference (Image generation):}
Once an optimal $z_T^G$ is found, generating an image is done by following the standard denoising process of the DDPM to obtain $I^{G}$. To generate multiple different images one can repeat the optimization by sampling a new $z_{T}^{G}$ and optimize it using \ourmethod{}. See a faster method below.
\newline

\subsection{Improving Speed and quality}
\label{sec:improving}

\noindent\textbf{Stabilized optimization.}
The last few denoising steps $z_{t}$,$z_{t-1}$, .. $z_{0}$ for $t<<T$, often generate high quality images. To speed up convergence, we compute the losses for all images in the last $t$ steps  $\mathcal{L}_{Semantic}^{t}$,  and then aggregate them  $\mathcal{L}_{Semantic} = \sum_{i=0}^{t}{\mathcal{L}_{Semantic}^{i}}$. In our experiments, we found $t=2$ to be suitable to stabilize optimization.

\noindent\textbf{Faster generation using bootstrap.}
Typically, finding an optimal $z_T^G$ takes between 1-4 minutes on an NVIDIA A100 GPU. To quickly generate a large number of images, we operate as follows. First,  execute the optimization procedure, with fewer iterations, to find an optimal $z_T^G$ for the full set $\mathcal{I} = \{I_1,\ldots, I_k\}$. Then, use bootstrap \cite{efron1992bootstrap} to sample a subset $S \subset \mathcal{I}$ of reference images. Finally, find an optimal $z^{G_{S}}_T$ for the subset $S$, but start the optimization from $z^G_T$ and generate the image $I^{G_S}$. We repeat this process for multiple subsets to obtain a  diverse set of images. We find this bootstrap procedure, where we first learn a good initialization point and then generate images based on subsets, reduces the optimization duration for a single image from minutes to seconds.

\noindent\textbf{Contrasting classes.}
When generating images from a set of classes $C$, we can further improve optimization convergence and image quality by using a supervised contrastive loss~\cite{khosla2020supervised}. The loss operates in the \textit{semantic space}; it  pulls
 the semantic vector $v^{G}$ closer to the centroid of its class $\mu_v^{c}$, and pushes it away from centroids of other classes $\mu_v^{c'}$. The updated semantic loss is
\begin{equation}
    \mathcal{L}_{Semantic} = -\log\frac{e^{-dist(\mu_{v}^{c},v^{G})}}{\sum_{c' \in C}{e^{-dist(\mu_{v}^{c'},v^{G})}}} \quad.
\end{equation}

\section{Experiments} \label{sec:experiments}

To assess the quality of images generated for rare concepts, we analyzed several important aspects: faithfulness (whether the correct concept was generated), visual appeal (realism and naturalism of the image), diversity of generated images, and applicability to downstream applications.

To evaluate the faithfulness of generated images, we used SoTA pre-trained classifiers to determine whether the images belong to the correct class or not, supplemented by human evaluations. For assessing image realism, we used the FID score to quantify the distinction between real and generated samples.

For diversity, we used standard measures to find the Precision,
Recall, Fidelity, and Diversity of generative models.

Finally, we assessed how \ourmethod{} can benefit two downstream applications: (1) for generating hand palm images, a challenging task since foundation diffusion models were published. and (2) as semantic data augmentations to enhance few-shot CLIP classification.

\subsection{Rare Concept Generation}
We evaluate the quality of images from rare classes generated by our approach.

\textbf{Datasets.} We evaluated \ourmethod{} on three common benchmarks: \textbf{(1) ImageNet \cite{ImageNet}:}, the cannonical dataset with 1000 classes. As shown in Figure 2, about 25\% of ImageNet classes are in the tail of Laion.
\textbf{(2) CUB \cite{CUB}:} A \textit{fine-grained} dataset with a total of 200 bird categories. Most of the classes are in the tail of the Laion distribution. \textbf{(3) iNaturalist~\cite{van2018inaturalist}}: A large-scale, \textit{fine-grained} dataset for species classification.
Its entire set of classes is in the tail of the Laion distribution.

We use CUB and iNaturalist since most of their classes are rare; i.e. have been represented by fewer than 10k samples in the training set of the diffusion model (more details in supp).

\textbf{Evaluation protocol.}
We ranked classes for each dataset according to their occurrence frequency in the LAION2B dataset. For each class, the set of reference images for all methods was taken from the trainset. Specifically, we sampled a maximum of 50 random images,  $k=\max{(|class|,50})$. Subsequently, we used different methods to generate images based on the real reference samples.

\textbf{Pretrained classifiers.} To measure the correctness of generated images, we use SoTA pre-trained classifiers  for each benchmark, sourced from open repositories available online. Specifically, for ImageNet, we used \cite{tu2022maxvit}, achieving 88.2\% accuracy on the corresponding test set. For CUB, we used \cite{Chou2023FinegrainedVC}, which attains 93.1\% test accuracy. For iNaturalist, we used  \cite{Ryali2023HieraAH}, which has 83.8\% test accuracy.

\begin{table}[t!]
\centering
    \begin{sc}
    \scalebox{0.95}{
    \setlength{\tabcolsep}{2pt} %
    \begin{tabular}{l|c}
     Method
     & FID $\downarrow$ \\
    \midrule
    SD \cite{StableDiffusion} & \textbf{6.4} \\
    SD+RG+RF \cite{He2022IsSD}& 6.9  \\
    Finetuned SD \cite{FTSD2023}& 10.2  \\
    \textbf{\ourmethod{} (ours)} &\textbf{ 6.5}  \\
    \bottomrule
    \end{tabular}
    }
    \end{sc}
    \captionof{table}
      {Quality of images generated using various approaches, measured using FID. Compared with SD, other methods for rare-concept generation hurt image quality, but \ourmethod{} maintains the same image quality as SD. \label{tab:FID_eval}}
\end{table}

\textbf{Compared Methods.}
We compared \ourmethod{} with the following methods. \textbf{SD ~\cite{StableDiffusion}:} Vanilla Stable Diffusion v2.1; \textbf{SD+RG+RF ~\cite{He2022IsSD}:}  Stable Diffusion v2.1  with Real Guidance (RG) and Real Filtering (RF).; and \textbf{Finetuned SD ~\cite{FTSD2023}:} Finetuning SD on real training samples for each class.
To ensure a fair comparison, we replicated the methods mentioned above with StableDiffusion v2.1 using the code published by the respective authors. This step was essential because the previous approaches relied on different versions of pre-trained text-to-image models, making it crucial to establish a consistent framework for evaluation.

\textbf{Generation Protocol:}
See supplementary material.

\begin{table}[t!]
\begin{subtable}{0.5\textwidth}
\sisetup{table-format=-1.2}   % 2 decimals, leave space for minus sign
\centering
    \scalebox{0.99}{
    \setlength{\tabcolsep}{4pt} %
    \begin{tabular}{l|ccc}
    Human Eval & \multicolumn{3}{c}{\textbf{ImageNet}} \\
    & Many & Med & Few  \\
    & \#\textgreater{}1M & 1M\textgreater{}\#\textgreater{}10K & 10K\textgreater{}\#    \\\hline
    Finetuned SD  & 48.01$\pm$1.01 & 41.55$\pm$1.55 & 15.84$\pm$2.29 \\
    \ourmethod{}  & \textbf{50.12$\pm$1.00} & \textbf{55.33$\pm$1.42} & \textbf{69.08$\pm$2.46} \\
    Neither      & 1.87$\pm$1.12 & 3.12$\pm$1.48 & 15.08$\pm$2.22  \\
    \bottomrule
    \end{tabular}
    }
\end{subtable}
\bigskip

\begin{subtable}{0.5\textwidth}
\sisetup{table-format=-1.2}   % 2 decimals, leave space for minus sign
\centering
    \scalebox{0.99}{
    \setlength{\tabcolsep}{5pt} %
    \begin{tabular}{l|c|c}
     Human Eval & \textbf{CUB} & \textbf{iNaturalist} \\
    & All & All \\ \hline
    Finetuned SD  & 20.18$\pm$2.31 & 14.45$\pm$2.77 \\
    \ourmethod{}  & \textbf{68.98$\pm$2.71} & \textbf{72.44$\pm$2.13} \\
    Neither   & 10.84$\pm$3.11 & 13.11$\pm$2.79 \\
    \bottomrule
    \end{tabular}
    }
\end{subtable}

\caption{Human evaluation for rare-concept generation. Values are the percentage of raters that selected each option.}
     \label{tab:user-study-rare}
\end{table}

\begin{figure*}[t!]
    \centering     \includegraphics[width=1.7\columnwidth]{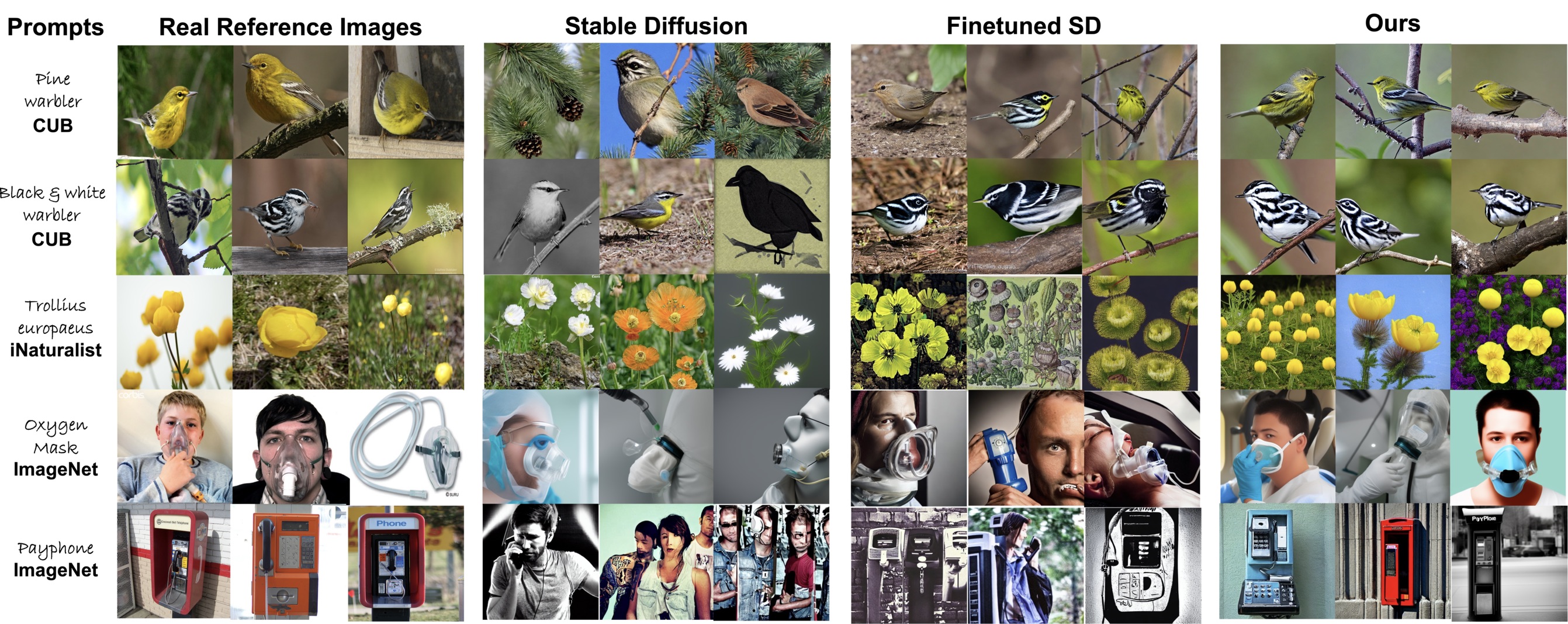}
    \caption{Qualitative comparison. Images generated by various methods for 5 rare classes from 3 datasets. Images generated by the competing techniques may exhibit high quality but frequently contain inaccuracies and fail to align with the real concept.
    }
    \label{fig:qualitative_classes}
\end{figure*}

\subsubsection{Evaluate faithfulness using pre-trained classifiers.}
Figure \ref{fig:rare_eval} shows per-class accuracy of different generation approaches on different benchmarks, as evaluated by pre-trained classifiers. Classes are ranked by their prevalence in the LAION2B dataset.  Notably, it shows that while existing methods falter in generating less common semantic concepts, our approach consistently achieves higher accuracy across all classes within all benchmarks.

\subsubsection{Evaluating realism and visual appeal.}
Table \ref{tab:FID_eval} further presents \ourmethod{} image quality in terms of realism and visual appeal compared to current generation methods. The measurement of this quality is done using the FID, which was calculated between 50K generated and 50K real ImageNet test images. The results demonstrate that SeedSelect's capability of generating rare concepts is not traded with image naturalism.
While other methods are negatively affected by adaptation to rare-concept generation, \ourmethod{} can generate images with the same quality as vanilla SD. This is attributed to the fact that these methods fine-tune the diffusion model or modify its denoising process, leading to a decline in image realism. In contrast, our method uses a pre-trained diffusion model with fixed parameters, only optimizing its seed during the generation process.

\subsubsection{Qualitative analysis.} Figure \ref{fig:qualitative_classes} compares images generated by Stable Diffusion and Finetuned SD \cite{FTSD2023} with our approach on rare concepts from CUB, ImageNet, and iNaturalist. See Suppl. for additional examples. The results show that although the compared methods generate realistic images of high quality they often fail to generate the correct concept.

\subsubsection{Evaluation with human raters.} We further performed a user study to analyze the correctness of the generated images. We randomly selected 30 classes from CUB, and iNaturalist, and 90 classes from ImageNet (30 for head, 30 for med, and 30 for tail). For each class we generated 10 images with \ourmethod{} and Finetuned SD \cite{FTSD2023}, the best baseline found in the previous analysis. Respondents were given the class name, three real samples as a reference, and the two generated images. They were asked to select which generated image better fits the class name and is semantically similar to the reference images. The final score for each approach is calculated as the number of times respondents selected the approach,  averaged across all the classes in the set. The study results are shown in Table \ref{tab:user-study-rare}. \ourmethod{} received the highest percentage of votes across all benchmarks: it is $\times3.4$ better on CUB and $\times5$ on the iNaturalist. Moreover, it excels across all splits of ImageNet with the most notable advancement observed within the tail, achieving a $\times4.3$ increase in accuracy. These results are correlated with the classifier results in Figure \ref{fig:rare_eval}.

\begin{table}[t!]
    \centering
      \begin{sc}
      \scalebox{0.85}{
    \setlength{\tabcolsep}{5pt} %
    \begin{tabular}{l|cccc}
     & SD & SD+RG+RF & Fintuned SD & Ours \\
    \midrule
    NDB $\downarrow$ & 2.48 & 2.6 & 2.9&2.52 \\
    Precision $\uparrow$& 0.79 & 0.70 & 0.61&0.77 \\
    Recall $\uparrow$& 0.2 & 0.15 &  0.11& 0.18 \\
    Fidelity $\uparrow$& 0.85 & 0.79&  0.71& 0.83 \\
    Diversity $\uparrow$& 0.37 &  0.28 & 0.20& 0.36 \\
    \bottomrule
    \end{tabular}
    }
    \end{sc}
    \caption{Diversity analysis comparison.
    \ourmethod{} generated samples with high diversity as SD,  while other approaches hurt diversity. }
  \label{tab:diversity_analysis}
\end{table}

\subsubsection{Diversity Analysis.}
A potential concern that may arise pertains to the diversity of the images generated by our approach. We analyze the diversity of images generated by \ourmethod{} compared to current methods using two measures of diversity. First, using NDB which finds diversity through mode collapse analysis,
\cite{richardson2018gans}.

Second, diversity as measured by \cite{naeem20a}, which directly assesses the coverage of generated samples compared to real samples. We also report Precision, Recall, and Fidelity.
These metrics were computed using 50K generated vs. 50K real ImageNet test images. We determined hyperparameters such as the number of clusters or neighbors, using the Elbow method \cite{thorndike1953belongs}.

The results of the diversity analysis are presented in Table \ref{tab:diversity_analysis}, indicating that \ourmethod{} maintains similar diversity to SD, whereas competing methods show lower diversity. We attribute this result to the fact that \ourmethod{} is initialized with a random seed and then optimized, leading to distinct images for each seed.

\subsubsection{Generation time.}
We compare our approach with personalized generation methods. Both Textual-Inversion~\cite{gal2022image} and  Dreembooth~\cite{ruiz2022dreambooth} require 30-60 minutes to learn a new single concept on a single NVIDIA A100 GPU. In contrast, \ourmethod{} with bootstrapping (See Section \ref{sec:improving}) takes 1-5 minutes to adapt to the new concept and 1-2 seconds to generate new semantically correct images.

\subsection{Hand generation}
As a first application, we test \ourmethod{} on hand generation.
Generating well-formed hand palms has been infamously hard to achieve with  diffusion models \cite{zhang2023adding}.
We tested how \ourmethod{} can be used for improving generation of hand palms.  Since there are currently no standard benchmark or automated methods to evaluate the quality of hand palm generation, we evaluated the results by asking human raters. In short, we used a 2-alternative-forced choice design (2AFC) asking raters to select if they prefer an image generated by \ourmethod{} or by SD.
The detailed procedure of the experiment is  in the supplemental material.

Table \ref{tab:user-study} shows the results of the user study. \ourmethod{} is $\sim\times4.5$ better in matching the prompt and $\sim\times4$ in generating realistic hands. More in supplementary.
Figure \ref{fig:qualitative_hands} compares Stable Diffusion with our approach on 5 hand prompts.

\subsection{Synthetic data for few-shot recognition}
\label{sec:exp_synthehic_data}
We  further examine the advantages of using \ourmethod{} for few-shot classification through semantic data augmentation.

In the context of few-shot image recognition, we are provided with a limited number of real training samples per class, along with their corresponding class names, and the goal is to fine-tune CLIP.

\textbf{Experimental Setup.} For a fair comparison we follow the same experimental protocol of \cite{Zhou2021LearningTP, Zhang2021TipAdapterTC}  and generation protocol of \cite{He2022IsSD}. Specifically, given a limited number of real training samples per class we generated 800 samples for each class using \ourmethod{}. We then fine-tuned a pre-trained CLIP-RN50 (ResNet-50). Fine-tuning is done using both real and generated images known as mix-training. More details on the setup/protocol can be found in \cite{He2022IsSD} and in Supp.

\begin{table}[t!]
    \centering
    \scalebox{1}{
    \setlength{\tabcolsep}{3pt} %
    \begin{tabular}{l ccc}
    & \multicolumn{3}{c}{\textbf{Rater decisions}}\\
     & Stable Diffusion & \ourmethod & Neither  \\ [0.5ex]
    \midrule
 Matches prompt  & 16.22$\pm$2.6 & \textbf{70.21$\pm$2.6} & 13.57$\pm$4.1  \\
    Looks realistic & 16.19$\pm$5.8 & \textbf{62.46$\pm$6.5} & 21.35$\pm$7.2  \\
    \bottomrule
    \end{tabular}
    }
    %\vspace{-5pt}
    \caption{Human evaluation of hand-palm generated images. Values are percentage of raters that selected each option. }
    \label{tab:user-study}
\end{table}

\begin{figure}[t!]
    \centering \includegraphics[width=0.99\columnwidth]{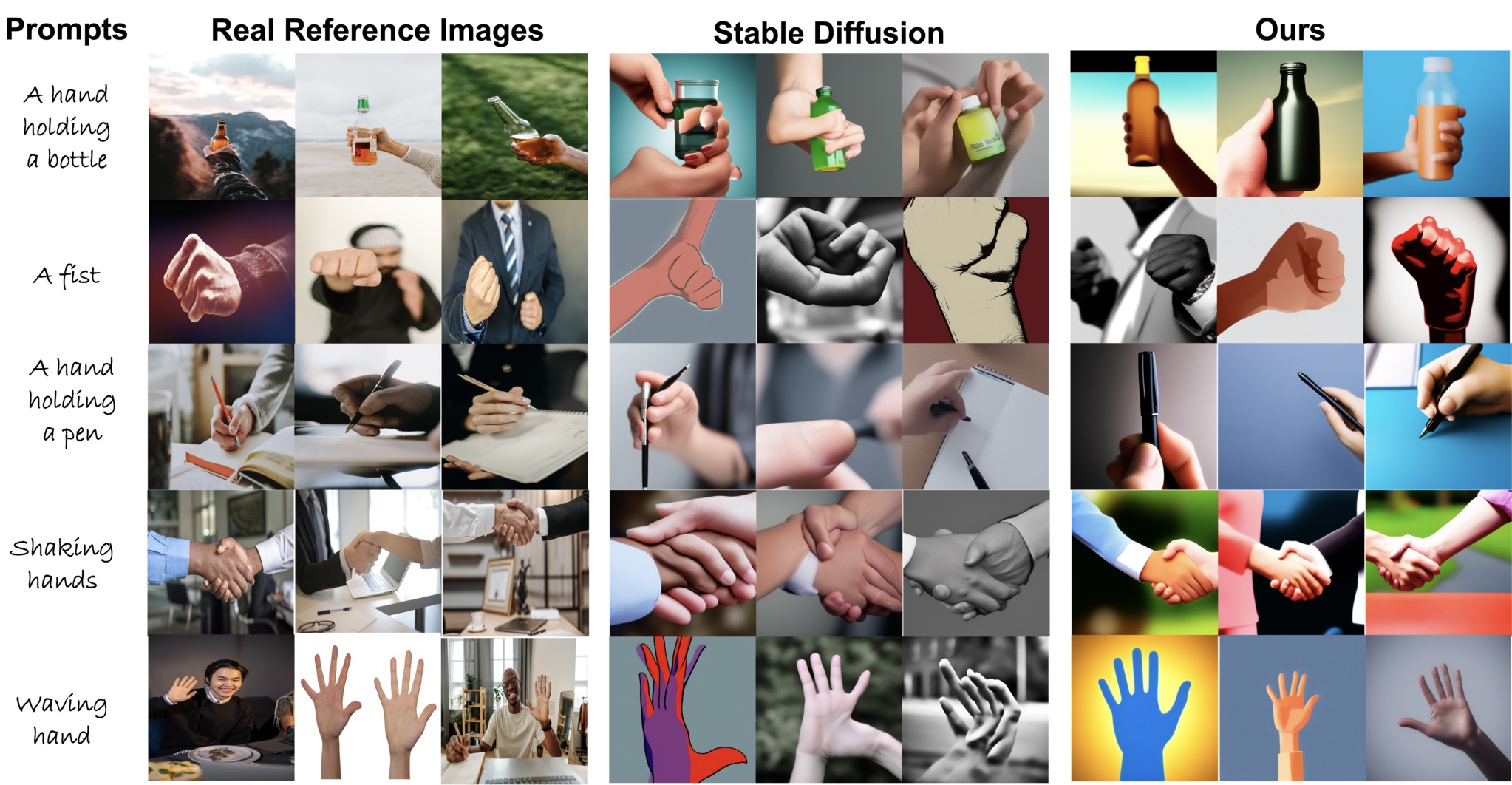}
    \caption{Qualitative comparison. Images generated by Stable Diffusion and \ourmethod{} for several hand generation prompts. While generated hands from SD are often corrupted \ourmethod{} can fix this shortcoming given a few reference examples.
    }
    \label{fig:qualitative_hands}
\end{figure}

\begin{figure}[t!]
    \centering
    \includegraphics[width=1\columnwidth]{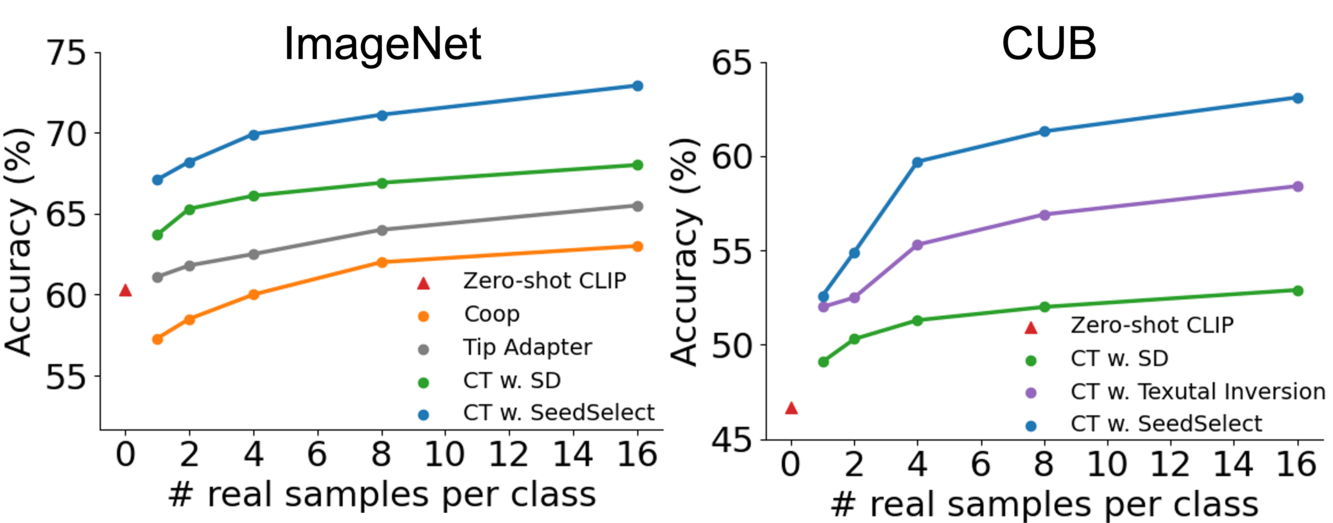}
    \caption{Results for few-shot image recognition, comparing \ourmethod{} to previous approaches.Fine-tuning a CLIP classifier on \ourmethod{} generated images consistently achieves SOTA results across all shot levels, with \ourmethod{} performing well even when given just a single image.
}
    \label{fig:fs_clip_eval}
\end{figure}

\textbf{Compared Methods.} We compare our approach with the following baselines: \textbf{Zeor-shot CLIP:} Applying the pre-trained CLIP classifier without fine-tuning; \textbf{CooP \cite{Zhou2021LearningTP}:}  Fine-tuning a pre-trained CLIP via learnable continuous tokens while keeping all model parameters fixed; \textbf{Tip Adapter \cite{Zhang2021TipAdapterTC}:} Fine-tuning
a lightweight residual feature adapter; \textbf{CT \& SD:} Classifier tuning with images generated with SD. \textbf{Textual Inversion \cite{gal2022image}:} Classifier tuning with images generated using personalized concepts (See Supp for implementation detail). Results for \textbf{CT \& SD} were reproduced by us on SDv2.1 using the code published by the respective authors.

\subsubsection{Results}
Figure \ref{fig:fs_clip_eval} shows results for the few-shot image recognition task. It demonstrates the effectiveness of \ourmethod{} in generating high-quality augmentations. When using \ourmethod{} augmentations to fine-tune a CLIP classifier, we achieve state-of-the-art performance across all shots. Remarkably, even with just a single image for training, \ourmethod{} still manages to produce valuable, diverse, and better augmentations compared to previous baselines. We provide results for iNaturalist in the Supp material.

\section{Discussion and limitations}

Although very powerful, modern text-to-image generation models still suffer from several shortcomings. They often generate incorrect images when prompted for rare concepts especially when a closely related concept appears frequently in the train set of the diffusion model.
We propose to remedy these issues by providing a handful of reference images of the concept to the diffusion model. Essentially, it selects a generation seed that drives the diffusion model to generate the correct concept, semantically and visually.
While \ourmethod{} offers a simple yet effective way to generate correct classes and hands, there are several limitations to consider. First, we find that it struggles with imitating the style of the reference images (e.g. when guided by sketch images of dogs, \ourmethod{} often generates natural images of dogs rather than sketches). Second, the optimized $z_T$ is prompt-specific, and doesn't generalize directly to other prompts.

Finally, for extremely rare concepts, with just a few examples in LAION2B, the quality of generated images is  poor.

\bibliography{aaai24}
\renewcommand{\figurename}{Fig. S}
\renewcommand\tablename{Table S}
\newcommand{\figref}[1]{S\ref{#1}}
\newcommand{\tabref}[1]{S\ref{#1}}
\newcommand{\cmark}{\ding{51}}%
\newcommand{\xmark}{\ding{55}}

\appendix

{\huge{\textbf{Supplemental Material}}}
\newline
\newline
We begin with a detailed explanation of computing concept distributions in LAION2B and LAIN400 (Appendix A) and then provide background on diffusion models (Appendix B).
We proceed to offer more details regarding the problem of rare-concept generation with a reference set (Appendix C), elaborate on the experimental protocol for few shot recognition (Appendix D), and provide implementation details (Appendix E). We further give more intuition and analysis of the semantic and natural-appearance losses (Appendix F and G), provide results for iNaturalist dataset (Appendix H), and present more information about our Textual Inversion baseline (Appendix I). Lastly, we provide supplementary qualitative results and more details about the user study done for hand generation (Appendix J and K).

\section{Computing Distribution of concepts in LAION}
\label{sec:supp_concept_dist}

We now describe, how we extracted concepts and class frequency from the large-scale LAION2B and LAION400.
First, we downloaded all text prompts from the official LAION website\footnote{https://laion.ai/blog/laion-5b/}. We then pre-processed each prompt to contain only alphabetic and numeric characters. We split each text-prompt to unigrams and bigrams. Finally, we counted the number of occurrences for each unique unigram and bigram.

Since LAION2B and LAION400 are large-scale datasets we used Apache Spark \cite{zaharia2016apache}. We ran the Spark application on an Amazon EMR cluster with 3 servers, each with 36 CPU cores and 60 GB of memory. It took about 4 hours to extract all phrases. Unigrams and bigrams will be published.

Figure \figref{fig:supp_laion2b} illustrates the distribution of concepts in LAION2B~\cite{Laion5B_dataset} and LAION400M~\cite{schuhmann2021laion}, introducing a long-tail behavior.

\section{Background and preliminaries}
Here we provide further information regarding Stable Diffusion~\cite{StableDiffusion}.

Stable diffusion operates in the latent space of a variational auto-encoder. First, an encoder $\mathcal{E}$ is trained to map a given image $x$ into a spatial latent code $z=\mathcal{E}(x)$. Subsequently, a decoder $\mathcal{D}$ is then tasked with reconstructing the input image such that $\mathcal{D}(\mathcal{E}(x)) \approx x$, thus ensuring that the latent representation accurately captures the original image.

Following the training of the autoencoder, a denoising diffusion probabilistic model (DDPM) \cite{ho2020denoising} operates over the learned latent space to produce a denoised version of an input latent $z_t$ at each timestep $t$. During the denoising process, the diffusion model can be conditioned on an additional input vector. In Stable Diffusion, this additional input is typically a text encoding produced by a pre-trained CLIP text encoder \cite{radford2021learningCLIP}. The conditioning vector is denoted as $c(y)$, given conditioning prompt $y$. Finally, the DDPM model $\epsilon_{\theta}$ is trained to minimize the loss:
\begin{equation}
      \mathcal{L} =
    \mathbb{E}_{z\sim\mathcal{E}(x),y,\varepsilon\sim\mathcal{N}(0,1),t}
    \left [ || \varepsilon - \varepsilon_\theta(z_t, t, c(y)) ||_2^2 \right ].
\end{equation}
In essence, the denoising network $\varepsilon_\theta$ is required for correctly removing noise $\varepsilon$ that has been added to the latent code $z$ at each time-step $t$, given the noised latent $z_t$ and conditioning encoding $c(y)$. $\varepsilon_\theta$ is a UNet network~\cite{ronneberger2015u} that incorporates self-attention and cross-attention layers.

At inference, a latent $z_T$ is sampled from a standard multivariate normal distribution $\mathcal{N}(0, 1)$ and is iteratively denoised using the DDPM to produce a latent $z_0$. This denoised latent is then passed to the decoder $\mathcal{D}$ to obtain the image $I^{G} = \mathcal{D}(z_0)$.

\section{Rare concept generation}
\begin{figure}[t!]
    \centering
    \includegraphics[width=1\columnwidth]{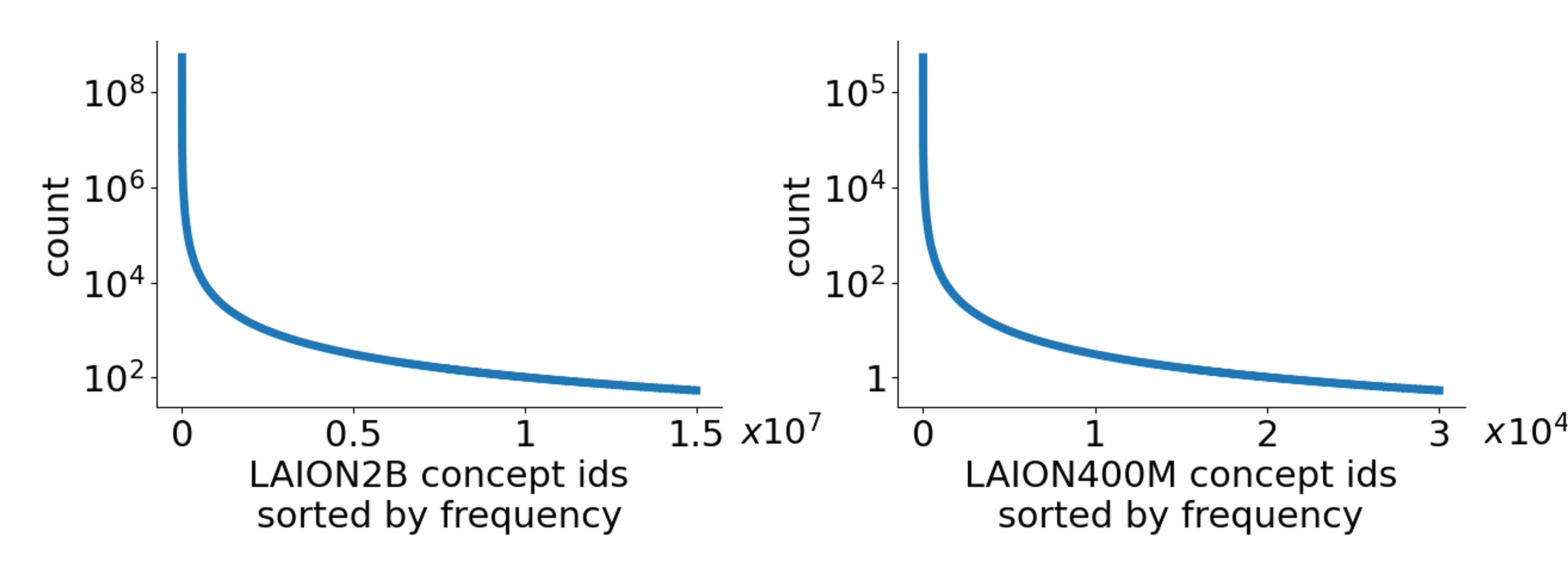}
    \caption{LAION2B and LAION400M concept ids sorted by frequency. A long-tail distribution is observed.}
    \label{fig:supp_laion2b}
\end{figure}

Here we describe the details of the evaluation done for images generated from rare concepts (in the long-tail of categories in Fig. \figref{fig:supp_laion2b}). First, we explain the generation protocol for evaluation, then give more details about the pre-trained classifier and how the classes were ranked. Finally, we provide additional results comparing per-class accuracy between \ourmethod{} and current generation methods.

\paragraph{Generation protocol:}
One weakness of diffusion models is their sensitivity to prompts. Different prompts with the same meaning might yield images with different quality or concepts. To create a more robust generation, \cite{radford2021learningCLIP} proposed to use multiple prompts. Specifically, instead of encoding a single prompt in clip space and inputting it to Stable Diffusion, one provides the centroid (mean) of all encoded prompts.
The centroid then guides the diffusion model generation. Section \ref{sec:supp_template_prompts} below lists the template of prompts used in our experiments.

\begin{figure}[t!]
    \centering
    \includegraphics[width=1\columnwidth]{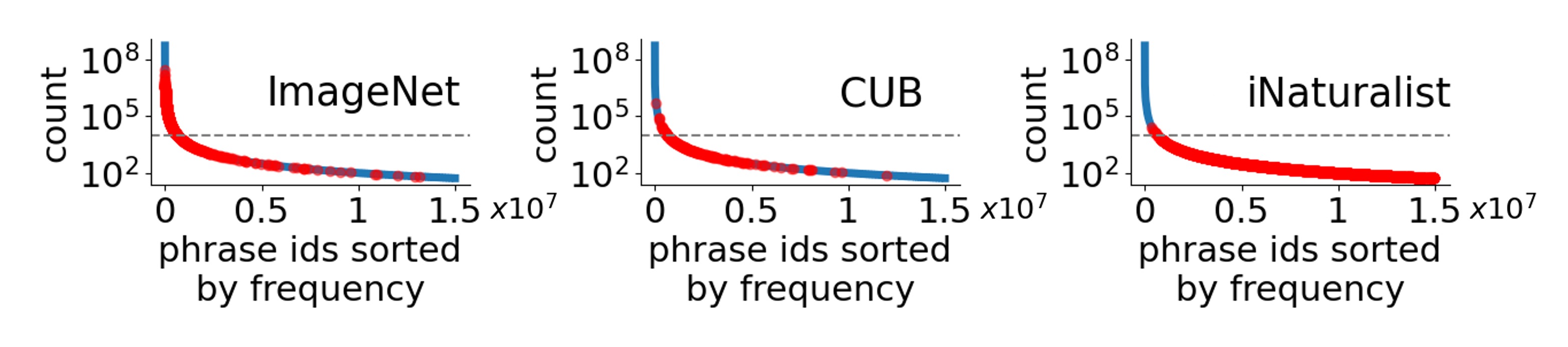}
    \caption{Class counts (red) on top of LAION2B phrase distribution (blue). It shows that most of the classes on these benchmarks are at the tail of LAION2B.  The horizontal line at $10^4$ counts shows the 25\% tail according to Fig. \ref{fig:laion}.}
    \label{fig:class_counts_laion}
\end{figure}

\begin{table}[t!]
\begin{subtable}{0.5\textwidth}
\sisetup{table-format=-1.2}   % 2 decimals, leave space for minus sign
\centering
   \scalebox{0.9}{
    \setlength{\tabcolsep}{3pt} %
    \begin{tabular}{l|cccc}
     & \multicolumn{4}{c}{\textbf{ImageNet1k in LAION2B}} \\
     \midrule
    Methods & Many & Med & Few & Total Acc  \\ [0.5ex]
    & n=235 & n=509 & n=256 &   \\ [0.5ex]
    & \#\textgreater{}1M & 1M\textgreater{}\#\textgreater{}10K & 10K\textgreater{}\# &   \\ [0.5ex]
    \midrule
    SD & 97.1 & 94.5 & 52.2 & 84.5 \\
    SD + RG & 96.9 & 94.5 & 52.2 & 84.5 \\
    Finetuned SD & 96.9 & 94.3 & 61.4 & 86.7 \\
    \midrule
    \ourmethod{} (\textbf{ours}) & \textbf{97.8} & \textbf{95.8} & \textbf{76.1} & \textbf{91.3} \\
    \bottomrule
    \end{tabular}
    }
\end{subtable}
\bigskip

\begin{subtable}{0.5\textwidth}
\sisetup{table-format=-1.2}   % 2 decimals, leave space for minus sign
\centering
   \scalebox{0.9}{
    \setlength{\tabcolsep}{3pt} %
    \begin{tabular}{l|cccc}
     & \multicolumn{4}{c}{\textbf{CUB in LAION2B}} \\
     \midrule
    Methods & Many & Med & Few & Total Acc  \\ [0.5ex]
    & n=0 & n=22 & n=178 &   \\ [0.5ex]
    & \#\textgreater{}1M & 1M\textgreater{}\#\textgreater{}10K & 10K\textgreater{}\# &   \\ [0.5ex]
    \midrule
    SD & - & 88.2 & 52.0 & 55.9 \\
    SD + RG & - & 89.6 & 55.1 & 58.8 \\
    Finetuned SD & - & 91.0 & 60.9 & 64.2 \\
    \midrule
    \ourmethod{} (\textbf{ours}) & - & \textbf{98.1} & \textbf{80.9} & \textbf{82.8} \\
    \bottomrule
    \end{tabular}
    }
\end{subtable}

\bigskip
\begin{subtable}{0.5\textwidth}
\sisetup{table-format=-1.2}   % 2 decimals, leave space for minus sign
\centering
   \scalebox{0.9}{
    \setlength{\tabcolsep}{3pt} %
    \begin{tabular}{l|cccc}
     & \multicolumn{4}{c}{\textbf{iNaturalist in LAION2B}} \\
     \midrule
    Methods & Many & Med & Few & Total Acc  \\ [0.5ex]
    & n=0 & n=7 & n=7070 &   \\ [0.5ex]
    & \#\textgreater{}1M & 1M\textgreater{}\#\textgreater{}10K & 10K\textgreater{}\# &   \\ [0.5ex]
    \midrule
    SD & - & 79.6 & 26.9 & 26.9 \\
    SD + RG & - & 86.9 & 29.1 & 29.2 \\
    Finetuned SD & - & 96.3 & 34.1 & 34.1 \\
    \midrule
    \ourmethod{} (\textbf{ours}) & - & \textbf{98.1} & \textbf{68.8} & \textbf{68.9} \\
    \bottomrule
    \end{tabular}
    }
\end{subtable}

\caption{Image generation quality measured by the accuracy of pre-trained classifiers. Average per-class accuracy is reported separately for head ("Many", classes with over 1M instances in LAION2B), tail ("Few", classes, with less than 10K instances), and middle ("Med", classes, the rest) classes. n is the number of classes at each split. \ourmethod{} achieves improvement in all classes, with particular significance on the tail, over all benchmarks.}
     \label{laion-imagenet-bench}
\end{table}

\paragraph{Class frequency and ranking: }
We ranked classes by their number of occurrences using unigrams and bigrams extracted from the LAION2B dataset (see section \ref{sec:supp_concept_dist}). Since stable diffusion is conditioned on features extracted from CLIP~\cite{radford2021learningCLIP}, the features of two synonyms are often close to each other in the clip space, generating two similar images. Therefore, rare concept might be close to another frequent concept in the clip space and the ranking would not represent the true distribution in LAION2B. To overcome this challenge we gathered all class synonyms, using the nltk library~\cite{bird2009natural}, and count them as well. The number of samples for that class is therefore the sum of occurrences of that class and all its synonyms.

\paragraph{Benchmark concept distribution:} Figure \figref{fig:class_counts_laion} traces the relative frequency of classes from CUB, iNaturalist, and ImageNet,
as found in the LAION2B dataset. It shows that datasets like
CUB and iNaturalist are almost entirely in the tail of the distribution (bottom 25\% of classes, dashed line).

\paragraph{Comparing with SoTA approaches:} We compared \ourmethod{} with two SoTA rare concept generation approaches: SD+RG+RF ~\cite{He2022IsSD} and Finetuned SD~\cite{FTSD2023}. To ensure a fair comparison, we used the code published by the authors and apply them to the StableDiffusion used by us (StableDiffusion v2.1). This step was essential because the previous approaches relied on different versions of pre-trained text-to-image models. This way, we established a consistent framework for evaluation.

Our evaluation framework aligned with that of the original authors, involving identical training and inference procedures. In the case of SD+RG+RF~\cite{He2022IsSD}, the process began with RealGuidance (RG), which introduced noise to the reference images over a set number of steps denoted as $t$. Subsequently, StableDiffusion was applied to restore clarity to the images, resulting in new images depicting the same concept. We adhered to the author's suggestion of setting $t$ to 15 steps. Finally, RealFiltering (RF) is applied to filter images that are similar to the reference images.

For the Finetuned SD method, we used the reference images provided to fine-tune the UNet module, while keeping the other modules unchanged. The fine-tuning procedure was set to 210K steps. During the generation phase, we adopted a guidance scale of 1.5, again in line with the recommendations of the original authors.

\begin{figure*}[t!]
    \centering
    \includegraphics[width=1.6\columnwidth]{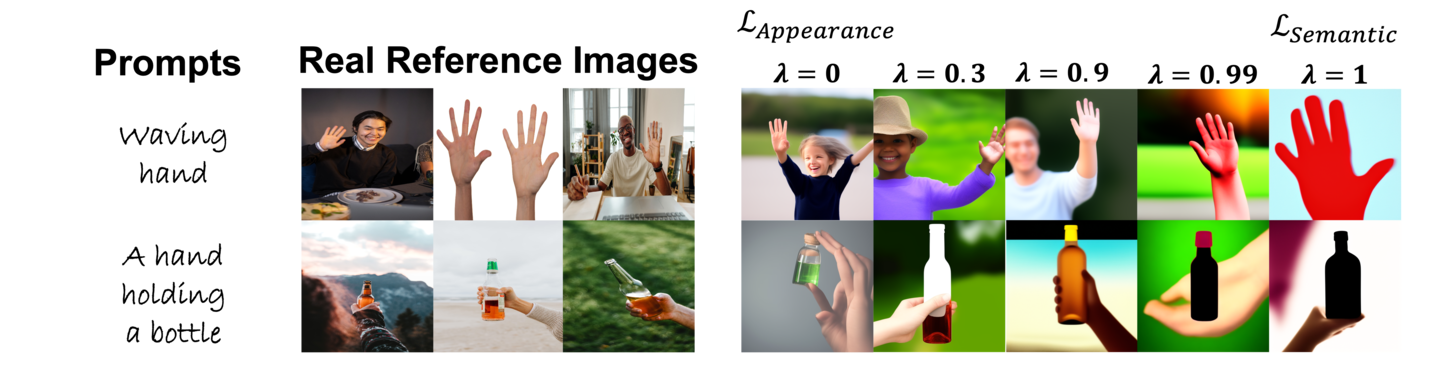}
    \caption{Effect of lambda on the generated image when starting from the same noise $z_T^G$ where the overall loss is: $\mathcal{L}_{Total} = \lambda \mathcal{L}_{Semantic} + (1-\lambda) \mathcal{L}_{Appearance}.$ Larger $\mathcal{L}_{appearance}$ generates more natural (low FID) and similar images to the few-shot training samples while larger $\mathcal{L}_{semantic}$ generates images that are semantically similar to the training samples.}
    \label{fig:ablation_lambda}
\end{figure*}

\paragraph{Evaluating faithfulness:} Table \tabref{laion-imagenet-bench} shows results from Fig. 4 in a tabular way and compares
 current methods and \ourmethod{} in terms of faithfulness and correctness. We report per-class accuracy for many-shot classes (appeared more than 1 million times in LAION2B), few-shot classes (appeared less than 10k times), and median-shot classes (the rest). \ourmethod{} improves per-class accuracy in all splits over all benchmarks.

\section{Experiment details for Synthetic Data for Few Shot Recognition}
Here, we provide experimental details on the experiments done in Section \ref{sec:exp_synthehic_data} of the main paper.

For few-shot image recognition, we adopt the experimental procedure outlined in \cite{He2022IsSD}. Specifically, we generate 800 images per class. Regarding Classifier Tuning (CT) methodologies, we adopt the "mix training" approach. Specifically, in each training iteration, we input a batch of real data into the model, calculate the loss for the real data portion, then similarly input a batch of synthetic data and compute the loss for the synthetic data segment. The final loss, obtained by summing these two loss values, is used for backpropagation and model fine-tuning.

\section{Implementation details and hyperparameters}

\paragraph{Generation with stable diffusion.} \ourmethod{} employs Stable Diffusion v2.1 with a guidance scale of 7.5 and 7 denoising steps using EulerDiscreteScheduler~\cite{karras2022elucidating}. The single hyper-parameter for our approach is $\lambda$, which controls the tradeoff between semantic loss and appearance loss. We determined the best $\lambda$ value by using the pre-trained ImageNet classifier accuracy as a measure to evaluate the quality of generated images. We found that $\lambda=0.9$ usually results in good-looking and semantically correct images and used the same value for all of our experiments. See Appendix \ref{sec:supp_lambda_effect} for the effect of different $\lambda$ values on the quality of generated images.

\paragraph{Few-shot recognition:} For few-shot recognition task, we used the AdamW optimizer, employing a weight decay of 0.1, and applying the cosine annealing rule. We set the batch size to 32 for few-shot real images and 512 for synthetic images. The training duration spans 30 epochs, with an initial learning rate of 0.001. Notably, the loss values from real data and synthetic data are equally weighted (1:1 ratio) in each iteration.

\begin{figure}
  \begin{minipage}[t]{0.5\linewidth}
    \centering     \includegraphics[width=1\columnwidth]{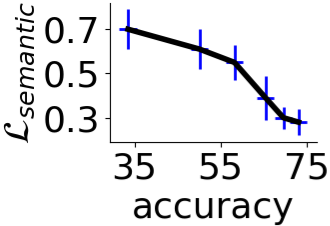}%
    \caption
      {%aaaa
      A clear relationship exists between the semantic loss and the semantic quality of a pre-trained classifier, underscoring the efficacy of this loss in optimization.
        \label{fig:semantic_loss_supp}%
      }%
  \end{minipage}\hfill
  \begin{minipage}{.4\linewidth}
    \centering
      \begin{small}\begin{sc}
      \scalebox{1}{
    \setlength{\tabcolsep}{2pt} %
    \begin{tabular}{l|cc}
    \textbf{\#shots} & \textbf{FID}\\
    \midrule
    0 & 7.8 \\
    5 & 6.5 \\
    10 & 6.45 \\
    20 & 6.1 \\
    60 & 6.0 \\
    100 & 5.8\\
    \bottomrule
    \end{tabular}
    }
    \end{sc}\end{small}
    \captionof{table}
      {%
        Image quality improves with more shots for $\mathcal{L}_{Appearance}$, showing that the loss is effective for natural apprearance consistency.
        \label{tab:FID_eval_appearance_loss}%
      }
  \end{minipage}
\end{figure}

\section{Semantic Loss and Appearance Loss}
Here we provide more intuition about the "semantic loss" and "natural appearance loss" used to train \ourmethod{}.

\paragraph{Semantic consistency:}
Our semantic loss is based on CLIP similarity, thus
reliability is reinforced by the fact that CLIP was trained on a vast amount of data, enabling it to capture plausible semantic estimates of the centroids of a small set of samples.
We have evaluated this relation quantitatively. We
trace the loss during optimization, and at the same time repeatedly compute a measure of semantic quality, by using ImageNet SoTA pre-trained classifier for the sake of analysis.
Figure \figref{fig:semantic_loss_supp} shows that the classification accuracy is correlated with our semantic loss along the seed optimization process.

\paragraph{Natural appearance consistency:}
Our appearance loss is the \textit{same loss} used in training the diffusion model: MSE between the latent tensor of the generated image and the latent of the given real images. It can be viewed as a way to keep natural consistency.
To show the effect of the appearance loss we have computed $\mathcal{L}_{appearance}$ using more images while keeping the semantic loss unchanged, and measuring FID (which is a standard metric to quantify image naturalism). Specifically, we fixed the $\lambda$ value and used five constant images for $\mathcal{L}{semantic}$. We then changed the number of reference images for $\mathcal{L}_{appearance}$  from 0 to 100. The imbalance in the final loss attributes any visual quality improvement to appearance loss. Table \tabref{tab:FID_eval_appearance_loss} shows that image quality has improved in terms of FID, showing the effectiveness of the appearance loss to keep the natural appearance consistency of the generated images.

\begin{table}[t!]
    \centering
      \begin{small}\begin{sc}
      \scalebox{1}{
    \setlength{\tabcolsep}{2pt} %
    \begin{tabular}{l|cc}
    \textbf{\#shots} & \textbf{FID}\\
    \midrule
    Zero-shot CLIP & 3.4 \\
    CT \& SD & 71.8$\pm$0.9 \\
    VL-LTR & \textbf{74.6} \\
    CT \& SeedSelect (\textbf{ours}) & \textbf{74.8$\pm$0.6} \\
    \bottomrule
    \end{tabular}
    }
    \end{sc}\end{small}
    \caption{Comparing \ourmethod{} on iNaturalist. Top-1 accuracy is reported. \ourmethod{} achieve comparable results to SoTA.}
  \label{tab:inaturalist_supp}
\end{table}

\section{Semantic vs Appearance tradeoff}
\label{sec:supp_lambda_effect}

Figure \figref{fig:ablation_lambda} studies the effect of changing the trade-off between semantic and appearance loss in the overall loss function:
\begin{equation}
    \mathcal{L}_{Total} = \lambda \mathcal{L}_{Semantic} + (1-\lambda) \mathcal{L}_{Appearance} \quad.
\end{equation}

It illustrates that larger $\lambda$ (i.e larger $\mathcal{L}_{Semantic}$) pushes the model to generate more feature-consistent images where smaller $\lambda$ values (i.e larger $\mathcal{L}_{Appearance}$)  pushes the model to generate more realistic samples to the given few-shot images.

\begin{figure}[t!]
\fbox{\begin{minipage}{23em}
\textit{
I have a photo that contains object for classification. I am using a model that generates images for classes that have few samples to increase the dataset.
\newline
\newline
Generate prompts that include the object. Use the format of: “A photo of 'OBJECT NAME' ..."
\newline
\newline
Be as diverse as possible, but be realistic about the scenes this object can appear in. \textbf{Be concise, but descriptive.} Include one prompt, in which the object appears without any background at all.
\newline
\newline
The number of prompts is 'NUMBER OF PROMPTS'.
The object is 'OBJECT'.}
\end{minipage}}
\caption{ ChatGPT prompt used to generate concise class prompts. The non-concise version is the same excluding the text in bold.}
\label{fig:chatgpt_prompts}
\end{figure}

\begin{figure}[t!]
    \centering     \includegraphics[width=0.9\columnwidth]{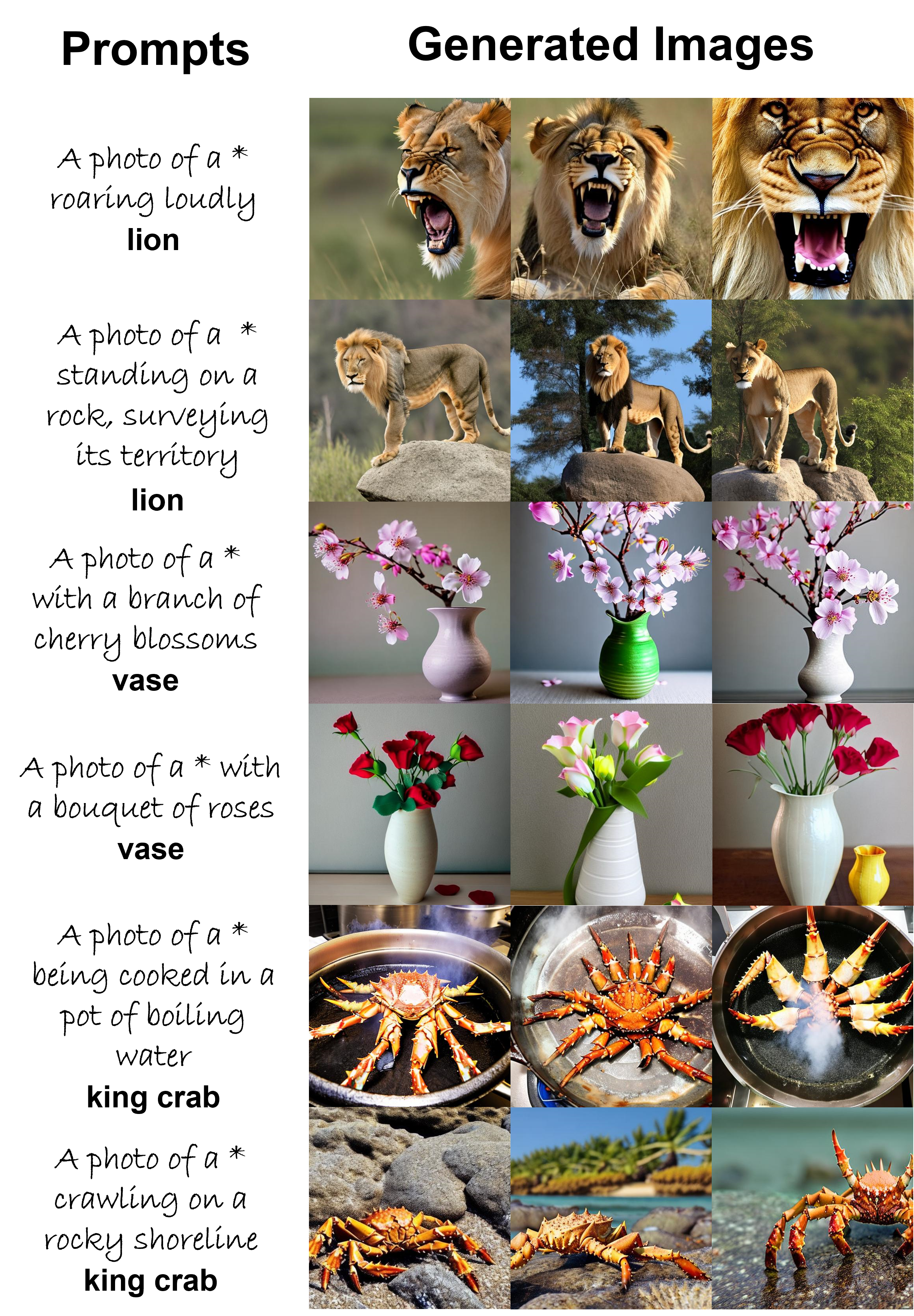}
    \caption{Images generated by Textual Inversion using ChatGPT prompts for several classes from ImageNet.
    }
    \label{fig:ti_qualitative_classes}
\end{figure}

\section{Results for iNaturalist}
Figure \ref{fig:fs_clip_eval} in the main paper compared our approach with common methods on the CUB~\cite{CUB} and ImageNet~\cite{ImageNet} datasets. We further provide results for the iNaturalist dataset. In this case, we used all training samples available in the train-set and generated samples for each class until the combined total of real and generated samples equaled the count of the class with the highest number of samples in the dataset, resulting in uniform data distribution. Subsequently, we fine-tune a CLIP classifier on both real and generated images. Table \figref{tab:inaturalist_supp} showcases how \ourmethod{} excels in producing high-quality augmentations, enabling fine-tuned CLIP classifier performance to achieve comparable results to SOTA methods tailored to the iNaturalist benchmark.

\section{Experiments with textual inversion}
\textbf{Textual Inversion~\cite{gal2022image}} is a recently proposed technique for personalization. In this paper, we adapted this method to generate "class concepts" images instead of specific "personalized concepts" and compared it with \ourmethod{} on CUB dataset~\cite{CUB} (Section \ref{sec:exp_synthehic_data} in the main paper).
Here, we provide more details about how we implemented it and show examples of generated images.

\begin{table}[t!]
    \centering
    \scalebox{0.9}{
    \setlength{\tabcolsep}{3pt} %
    \begin{tabular}{l|ccc|ccc}
     & \multicolumn{3}{c}{\textbf{Matches Prompt}} & \multicolumn{3}{c}{\textbf{Looks realistic}}\\
     \textbf{Prompts} & SD & Ours & Neither & SD & Ours & Neither\\ [0.5ex]
    \midrule
    \textit{\thead{"Shaking hands"}} & 12.12 & \textbf{71.63} & 16.25 & 12.12 & \textbf{75.16} & 12.72 \\
    \textit{\thead{"A hand holding \\ a bottle"}}& 20.11 & \textbf{73.21} & 6.68 & 17.91 & \textbf{61.22} & 20.87 \\
\textit{\thead{"A fist"}} & 16.22 & \textbf{70.97} & 12.81 & 10.11 & \textbf{56.65} & 33.24 \\
\textit{\thead{"A hand holding \\ a pen"}} & 17.12 & \textbf{69.99} & 12.89 & 26.65 & \textbf{58.49} & 14.84 \\
\textit{\thead{"Waving hand"}} & 15.52 & \textbf{65.27} & 19.2 & 14.19 & \textbf{60.72} & 25.09\\
    \bottomrule
    \end{tabular}
    }
    \caption{Human evaluation of hand-palm generated images. Values are the percentage of raters that selected each option. SD stands for Stable Diffusion. Note the significantly higher votes for our method.}
    \label{tab:supp_user-study}
\end{table}

\paragraph{Class concept generation:}
To generate diverse images for a given class we run the following procedure.

First, we used ChatGPT, a large language model based on InstructGPT~\cite{ouyang2022training} that was trained to engage in natural language conversations on a wide range of topics. We instructed ChatGPT to generate a list of 30 diverse class descriptions: 15 long descriptions and 15 concise but descriptive ones. We asked it to be as diverse as possible, but be realistic about the scenes the object of that class can appear in. See Figure \figref{fig:chatgpt_prompts} for the instruction given to ChatGPT.

\begin{figure*}[t!]
    \centering     \includegraphics[width=1.9\columnwidth]{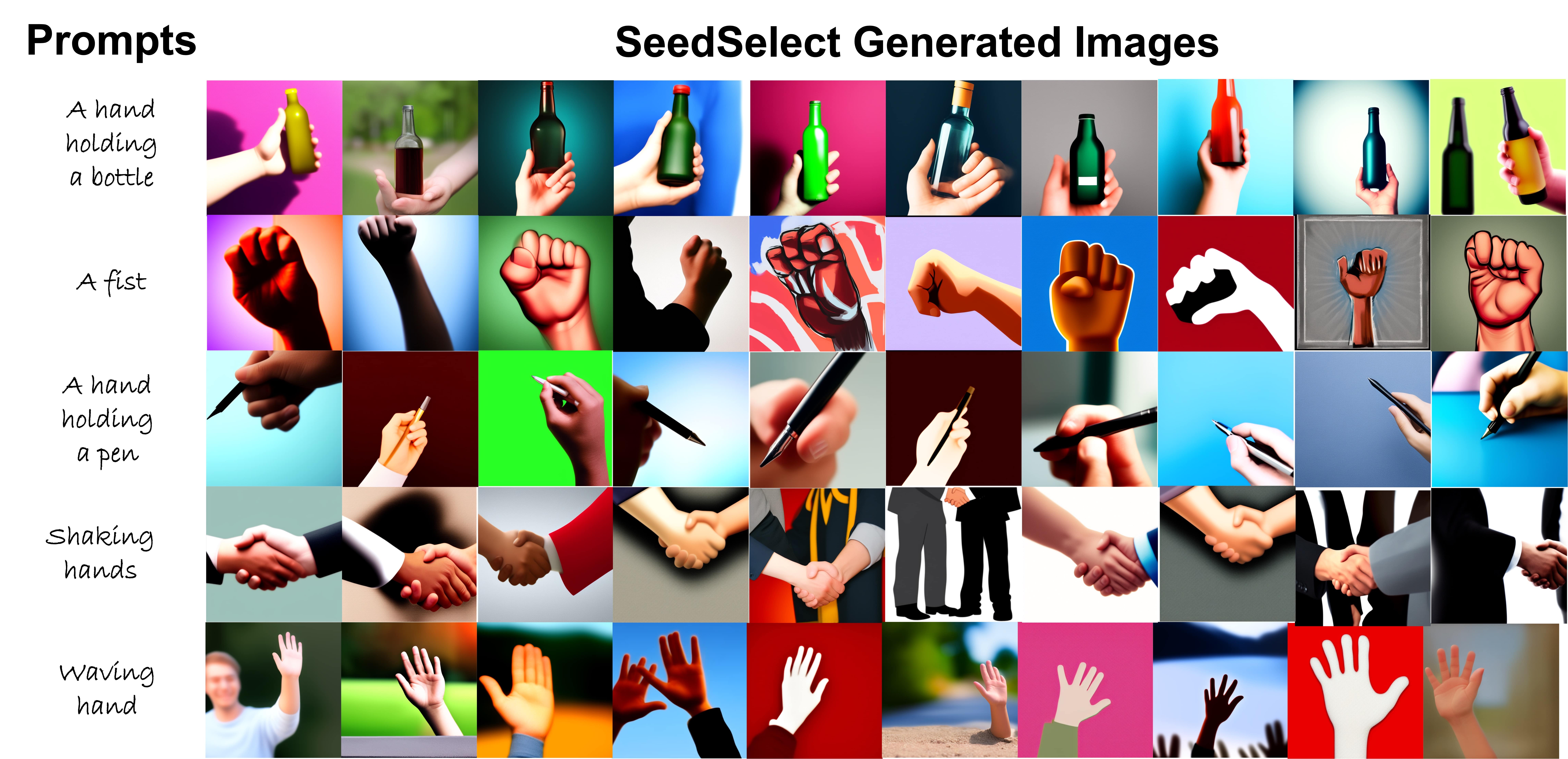}
    \caption{Uncurated images generated by \ourmethod{} for several hand generation prompts.
    }
    \label{fig:supp_qualitative_hands}
\end{figure*}

\begin{figure*}[t!]
\centering
    \includegraphics[width=2\columnwidth]{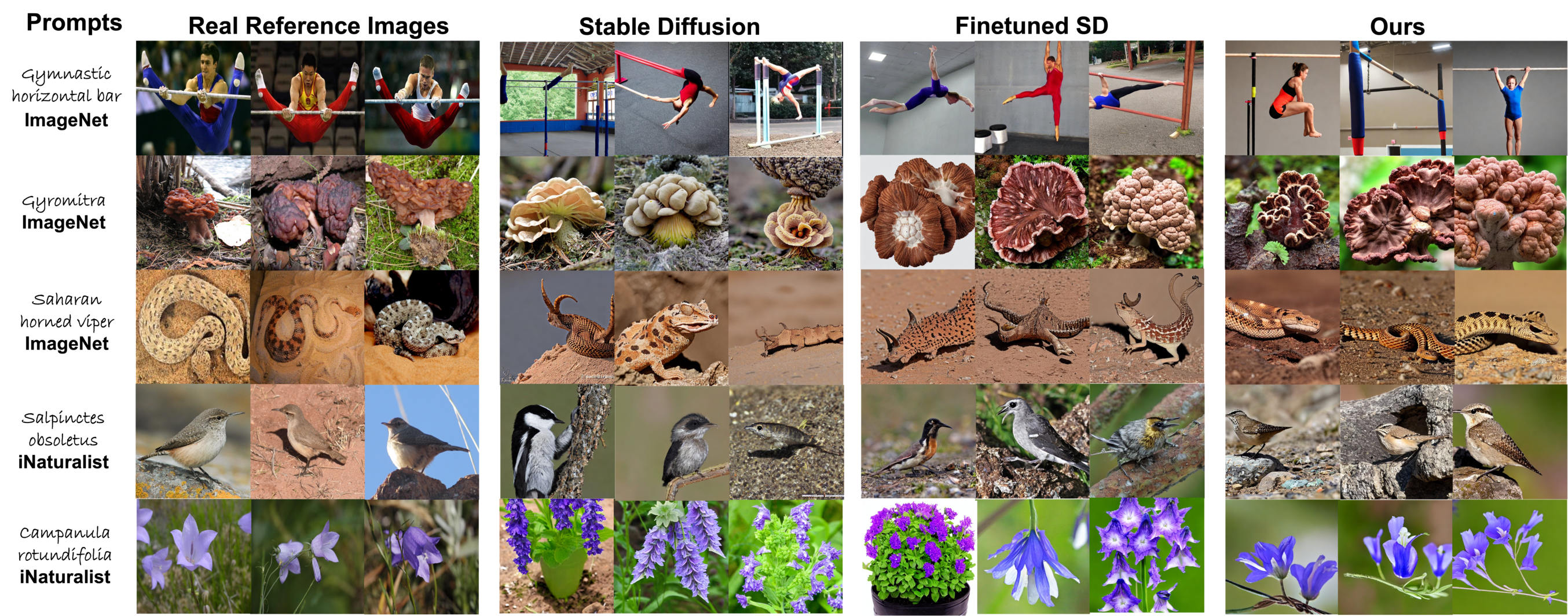}
    \caption{Additional qualitative comparison. Images generated by Stable Diffusion~\cite{StableDiffusion}, Finetuned SD~\cite{FTSD2023} and \ourmethod{} for several classes from ImageNet and iNaturalist.
    }
    \label{fig:supp_qualitative_classes}
\end{figure*}

Then, we trained Textual-Inversion on few-class images with the prompts obtained from ChatGPT and 80 short prompts proposed by \cite{radford2021learningCLIP}. During training, we resampled ChatGPT prompts x3 more than \cite{radford2021learningCLIP} to prevent bias towards short and less descriptive prompts. After training, we obtain an embedding token vector for the class.

Finally, we generated diverse images using the embedding vector and the list of prompts generated with ChatGPT.
Several generated samples can be found in Figure \figref{fig:ti_qualitative_classes}.

Note that since Textual-Inversion requires substantial computational resources when evaluating it as data augmentation
methods, we could only evaluate it on the CUB dataset, one of the few-shot recognition benchmarks.

\paragraph{Implementation Details:}
We used Stable Diffusion v2.1 as a base model. We train the embedding concept vector for 1000 steps. We used 50 denoising sampling steps and a guidance scale equal to 10.

\section{Experiments on Few-shot and Long-tail learning}
%\nir{ "and Results" or just "Results"}}
We further evaluate \ourmethod{} to improve generation quality in (1) semantic data augmentation to help improve few-shot classification, and (2) long-tail classification.

\subsection{Few-shot learning} \label{sec:fsl_experiments}
We test the benefit of using \ourmethod{} for few-shot classification by semantic data augmentation. In few-shot benchmarks, one is given a handful of samples per-class and their class names.
We use them to generate many more samples from that class, which are then used for training a classifier.
\newline

\noindent\textbf{Datasets.}
We evaluated \ourmethod{} on three common few-shot classification benchmarks:
\textbf{(1) CUB-200~\cite{CUB}:} A \textit{fine-grained} dataset consisting of 11,788 images of 200 bird species. Classes are split 100/50/50  meta-train/meta-val/meta-test.
\textbf{(2) miniImageNet~\cite{MatchNet}:} Derived from the standard  ImageNet dataset \cite{ImageNet},  with 64 classes for meta-training, 16 classes for meta-validation and 20 classes for meta-testing. Consists of 50000 training images and 10000 testing images, evenly distributed across all 100 classes.
\textbf{(3) CIFAR-FS~\cite{bertinetto2018cifarFS}:} Obtained from CIFAR-100 dataset \cite{CIFAR100} using the same criteria used for sampling miniImageNet. Containing 64 classes for meta-training, 16 classes for meta-validation, and 20 classes for meta-testing. Each class contains 600 images.
Following all previous baselines, we report classification accuracy as the metric. We report our results with 95\% confidence intervals on the meta-testing split of the dataset.
%\gal{Describe explicitely the split protocol }
% Fig. \ref{fig:class_counts_laion}
% %in section \ref{sec:data_overlap},
% provides, for each class in these datasets, its frequency in the LAION2B. It shows that miniImageNet has more classes in the head of LAION2B distribution and that in CUB, almost all classes are at the tail.

\noindent\textbf{Compared Methods.}
We compared our approach to recent few-shot SOTA methods. There are three main types of approaches.
(A) Methods that do not use pretraining nor use class labels for training: \textbf{Label-Hallucination ~\cite{Jian2022LabelHalluc}} and \textbf{FeLMi~\cite{roy2022felmi}}. (B) Methods that use class labels as additional information during training: \textbf{SEGA~\cite{yang2022sega}}. (C) Methods that use a classifier pre-trained on external datasets and use class labels as additional information during training: \textbf{SVAE~\cite{xu2022svae}}, \textbf{Vanilla Stable Diffusion (Version 2.1)~\cite{StableDiffusion}}, \textbf{Textual Inversion~\cite{gal2022image}} and \textbf{DiffAlign~\cite{DiffAlign}}.
The last two methods are semantic augmentations methods. As a result of the substantial computational resources required for \textbf{Textual Inversion}, we conducted our testing exclusively on miniImageNet (Details in supplementary).

~\newline\noindent\textbf{Experimental Protocol.}
For a fair comparison, we follow the training protocol of \cite{DiffAlign}. For each novel class, we generate additional 1000 samples using \ourmethod{} given the few-shot images provided during meta-testing and prompt with its class name. To perform N-way classification, we used a ResNet-12 model and trained it with cross-entropy loss on real and synthetic data.
To use also base classes information with \ourmethod, we used the contrasting classes technique (Sec. \ref{sec:improving}) and denoted it by \ourmethod+contrastive.

\begin{table*}[!t]
    \begin{minipage}{.3\linewidth}
    \centering
    \scalebox{0.65}{
    \begin{tabular}{lHH|cc|cc}
  %\toprule
  {\textbf{(a)}} & {} & {} & \multicolumn{2}{c|}{{\bf miniImageNet 5-way}} & \multicolumn{2}{c}{{\bf CIFAR-FS 5-way}} \\
  {\bf Model} & {{\thead{\bf  Vision \\ \bf \& Language}}} & {\bf Pretrained} & {\bf 1-shot} & {\bf 5-shot} & {\bf 1-shot} & {\bf 5-shot}\\
  \midrule
  {Label-Halluc ~\cite{Jian2022LabelHalluc}(AAAI'22)}  & {\xmark}&{\xmark}& { $67.04 \pm 0.7$} &{ $67.04 \pm 0.7$} & {78.03$\pm$1.0} & {89.37$\pm$0.6}\\
    {FeLMi~\cite{roy2022felmi} (NeurIPS'22)}  & {\xmark}&{\xmark}& { 67.47$\pm$0.8 } & { 86.08$\pm$0.4 }  &  { 78.22$\pm$0.7} & { 89.47$\pm$0.5}\\
    \midrule
    {SEGA~\cite{yang2022sega} (WACV'22) $\dagger$}  & {\cmark}&{\xmark}& { 69.04$\pm$0.3 } & { 79.03$\pm$0.2 }  & { 78.45$\pm$0.2} & {86.00$\pm$0.2} \\
    \midrule
    {SVAE~\cite{xu2022svae} (CVPR'22) $\ddagger$}  & {\cmark}&{\cmark}& {72.79$\pm$0.2} & { 80.70$\pm$0.2 }  & {73.25$\pm$0.4} & {78.89$\pm$0.3} \\
    {Textual Inversion~\cite{gal2022image}* $\ddagger$}  & {\cmark}&{\cmark}&
    {79.02$\pm$4.1} & {  85.44$\pm$3.9 }  & { -} & {-} \\
    {Stable Diffusion~\cite{StableDiffusion} $\ddagger$}  & {\cmark}&{\cmark}&
    {80.92$\pm$0.7} & {  85.05$\pm$0.5 }  & { 86.33$\pm$0.8} & {90.87$\pm$0.5} \\
  {DiffAlign~\cite{DiffAlign} $\ddagger$}  & {\cmark}&{\cmark}& {82.81$\pm$0.8} & { 88.63$\pm$0.3 }  & {88.99$\pm$0.8 } & {91.96$\pm$0.5} \\
  \midrule
  \textbf{\ourmethod{} (ours)}  & {\cmark}&{\cmark}& {\textbf{86.01$\pm$0.7}} & { \textbf{91.01$\pm$0.8} }  & {\textbf{91.12$\pm$0.5} } & {\textbf{93.46$\pm$0.4}} \\
  \textbf{\ourmethod{}+contrastive (ours) }  & {\cmark}&{\cmark}& {\textbf{87.02$\pm$0.8}} & { \textbf{92.08$\pm$0.7} }  & {\textbf{93.43$\pm$0.4} } & {\textbf{94.87$\pm$0.4}} \\
  \bottomrule
  \end{tabular}}
    \end{minipage}\hfill
    \begin{minipage}{.39\linewidth}
    \centering
    \scalebox{0.65}{
	\begin{tabular}{lHH|cc}
	%\toprule
	\textbf{(b)} & {} & {} & \multicolumn{2}{c}{\textbf{CUB 5-way}}    \\
	\textbf{Model} & {{\thead{\bf Vision \&
        \\ \bf Language}}} &
        {{\thead{\bf Pre \\ \bf trained}}} & \textbf{1-shot}     & \textbf{5-shot} \\
        \midrule
	TriNet  \cite{Chen2018MultiLevelSF} (TIP'19) & {\xmark}&{\xmark}&  69.61$\pm$0.5 & 84.10$\pm$0.4   \\
	FEAT  \cite{Ye_2020_CVPR} (CVPR'20)& {\xmark}&{\xmark}&  68.87$\pm$0.2 & 82.90$\pm$0.2   \\
	DeepEMD  \cite{Zhang_2020_CVPR}(CVPR'20) & {\xmark}&{\xmark}&  75.65$\pm$0.8 & 88.69$\pm$0.5   \\
        \midrule
        MultiSem  \cite{schwartz2022baby} (CoRR'19) $\dagger$ & {\cmark}&{\xmark}&  76.1$\pm$n/a & 82.9$\pm$n/a   \\
	SEGA~\cite{yang2022sega}(WACV'22) $\dagger$  & {\cmark}&{\xmark}& 84.57$\pm$0.2 & 90.85$\pm$0.2 \\
        \midrule
        Stable Diffusion* ~\cite{StableDiffusion} $\ddagger$ & {\cmark}&{\cmark}& 80.82$\pm$0.4 & 90.61$\pm$0.5 \\ \midrule
        \textbf{\ourmethod{} (ours)}  & {\cmark}&{\cmark}& \textbf{92.25$\pm$0.4} & \textbf{95.68$\pm$0.3} \\
        \textbf{\ourmethod{}+contrastive (ours)}  & {\cmark}&{\cmark}& \textbf{92.55$\pm$0.3} & \textbf{96.01$\pm$0.4} \\
        \bottomrule
	\end{tabular}
	}
    \end{minipage}
    \vspace{-5pt}
    \caption{\textbf{Few-shot recognition}. Comparison of \ourmethod{} to prior work on few-shot learning benchmarks. We report our results with 95\% confidence intervals on meta-testing split of the dataset. Our approach achieves the best results on all benchmarks. * denote reproduced by us. $\dagger$ for multi-modal methods that use class labels as additional information and $\ddagger$ for multi-modal methods that were also pre-trained on external datasets.}
    \label{table:few-shot-bench}
\end{table*}

~\newline\noindent\textbf{Results.}
Tables S\ref{table:few-shot-bench}a and  S\ref{table:few-shot-bench}b compare \ourmethod{}
with SoTA approaches for few-shot classification and semantic data augmentation, for  CUB, miniImageNet, and CIFAR-FS.

%The comparisons show that
\ourmethod{} consistently outperforms all few-shot benchmarks. The accuracy gain is largest in CUB, a fine-grained dataset whose classes are at the tail of the LAION2B training distribution, cutting down the 5-shot error from $9.39$ to $3.99$. This highlights the effectiveness of \ourmethod{} for rare and fine-grained class generation.

\subsection{Long-tail learning }
\paragraph{Datasets.}
We further evaluated \ourmethod{} in a task of long-tailed recognition using two major benchmarks.
\textbf{(1) ImageNet-LT~\cite{liu2019large}} A long-tailed version of the ImageNet dataset~\cite{deng2009imagenet} created by sampling a subset following the Pareto distribution with power value $\alpha = 6$. Consists of 115.8K images from 1000 categories with 1280 to 5 images per class.
Since class sampling is synthetic, classes that are at the tail of ImageNet-LT are not necessarily at the tail of LAION2B.
\textbf{(2) iNaturalist~\cite{van2018inaturalist}}: A large-scale, \textit{fine-grained} dataset for species classification. It is long-tailed by nature, with an extremely unbalanced label distribution with 437.5K images from 8,142 categories.
\newline\noindent\textbf{Compared Methods.} We compared our approach with three types of methods.
(A) Long-tail learning methods that do not use any pretraining nor employ class labels for training: \textbf{CE} (naive training with cross-entropy loss),  \textbf{MetaSAug~\cite{Li2021MetaSAugMS}}, \textbf{smDragon~\cite{samuel2020longtail}}, \textbf{CB LWS~\cite{Kang2019DecouplingRA}}, \textbf{DRO-LT~\cite{Samuel2021DistributionalRL}}, \textbf{Ride~\cite{ride}} and \textbf{Paco~\cite{Cui2021ParametricCL}}. (B) Methods that use class labels as additional information during training: \textbf{DRAGON~\cite{samuel2020longtail}}. (C) Methods that were pre-trained on external datasets and use class labels as additional information during training: \textbf{VL-LTR~\cite{Tian2021VLLTRLC}} and  \textbf{Vanilla Stable Diffusion (Version 2.1)~\cite{StableDiffusion}}.
\textbf{MetaSAug~\cite{Li2021MetaSAugMS}} and \textbf{Vanilla Stable Diffusion} are semantic augmentations methods. Note that \textbf{VL-LTR~\cite{Tian2021VLLTRLC}}, compared to other models,  further fine-tuned the pre-trained model (CLIP \cite{radford2021learningCLIP}) on the training sets. For \ourmethod, to use all training data information, we used the contrasting classes technique (see Section \ref{sec:improving}) and denoted it by \NSO{} + contrastive.
\newline

\begin{table}
\centering
      \scalebox{0.7}{
    \setlength{\tabcolsep}{5pt}
    \begin{tabular}{l|cc}
    % \multicolumn{1}{|r|}{Dataset}
    {Model} & \textbf{ImageNet-LT} & \textbf{iNaturalist}\\
    \midrule
    CE & 41.6 & 61.7   \\
    %LDAM Loss~\cite{cao2019learning} & - &68.0  \\
    MetaSAug~\cite{Li2021MetaSAugMS} & 47.4 &68.8 \\
    smDragon~\cite{samuel2020longtail} & 47.4 & 69.1 \\
    %$\tau-$norm~\cite{Kang2019DecouplingRA} & 46.7 &69.5 \\
    CB LWS~\cite{Kang2019DecouplingRA} & 47.7 &69.5   \\
    DRO-LT~\cite{Samuel2021DistributionalRL} & 53.5 &69.7 \\
    Ride~\cite{ride} & 55.4 &72.6 \\
    PaCO~\cite{Cui2021ParametricCL} & 53.5 & - \\
    \midrule
    DRAGON~\cite{samuel2020longtail} $\dagger$ & 57.0 & - \\
    \midrule
    VL-LTR~\cite{Tian2021VLLTRLC} $!$ & 70.1 &74.6 \\
    \midrule
    Stable Diffusion* $\ddagger$ & 56.4 &65.8 \\

\midrule
    \textbf{\ourmethod{} (ours)} & \textbf{73.5$\pm$0.3} & \textbf{74.5$\pm$0.9} \\
    \textbf{\ourmethod{}+contrastive (ours)} & \textbf{74.9$\pm$0.5} & \textbf{74.8$\pm$0.7} \\
    \bottomrule
    \end{tabular}}
    \vspace{-5pt}
\caption{\textbf{Long-tail recognition.} Values are accuracy, obtained with a ResNet-50 backbone. * denote reproduced by us. $\dagger$ for multi-modal methods that use class labels as additional information, $\ddagger$ for multi-modal methods that were also pre-trained on external datasets, and $!$ for multi-modal methods that further finetuned foundation models.
}
\label{table:long-tail-bench}
\end{table}

~\newline\noindent\textbf{Backbone.} Following previous methods, we use a ResNet-50 model architecture, train it on real and generated data, and report the top-1 accuracy over all classes on class-balanced test sets.
\newline

\noindent\textbf{Results.} Table S\ref{table:long-tail-bench} evaluates our approach compared to long-tail recognition benchmarks. It shows that by generating good semantic data augmentations, our approach achieves SoTA on both benchmarks compared to complex methods.

\section{Hand generation}
Table \ref{tab:user-study} in the main paper presents results for a user study done for hand generation. Here we give more details about the experiment and provide  results for each prompt separately.

Specifically, we collected from a popular Reddit forum\footnote{https://www.reddit.com/r/StableDiffusion}, the 5 most common prompts that stable diffusion fails to generate. These were:  "Shaking Hands", "Waving Hand", "A fist", "A hand holding a bottle" and "A hand holding a pen". We then collected 5 real "free-to-use" images from the internet that match these prompts. Then, we generated 150 images using Stable Diffusion (30 per prompt) and 150 images using \ourmethod{} trained with the 5 web images per-prompt. Finally, we constructed pairs of images (without cherry picking) and asked human raters to choose which image (1) matches the given description better and (2) shows a better well-formed hand.

Table \tabref{tab:supp_user-study} provides more results for each of the 5 hand prompts that were used in the user study. Showing that our method excels on all prompts, compared to Stable Diffusion.
See Figure \ref{fig:qualitative_hands} and Appendix \ref{sec:supp_qualitative} for qualitative results.

\paragraph{Stopping Criteria:} For hand generation, we utilized LPIPS Loss \cite{zhang2018unreasonable} to determine the stopping criteria for \ourmethod{}.We stop optimizing when the loss plateaus or its value increases for more than 3 iterations.

\section{Additional qualitative results}
\label{sec:supp_qualitative}
Figure \figref{fig:supp_qualitative_hands} provides uncurated images generated by \ourmethod{}.
Figure \figref{fig:supp_qualitative_classes} provides qualitative analysis and compares Stable Diffusion and Finetuned SD~\cite{FTSD2023} with \ourmethod{} on different classes from ImageNet and iNaturalist. The ``Gymnastic Horizontal Bar" demonstrates the capability of \ourmethod{} in creating a correct representation of the scene with reliable diversity, not necessarily visually similar to the provided real images.

\section{Prompts template}
\label{sec:supp_template_prompts}
Below we provide the list of prompt templates used for generating class images in \ourmethod{} method. * is replaced with the class name.

"a bad photo of a *", "a photo of many *", "a sculpture of a *", "a photo of the hard to see *", "a low resolution photo of the *", "a rendering of a *", "graffiti of a *", "a bad photo of the *", "a cropped photo of the *", "a tattoo of a *", "the embroidered *", "a photo of a hard to see *", "a bright photo of a *", "a photo of a clean *", "a photo of a dirty *", "a dark photo of the *", "a drawing of a *", "a photo of my *", "the plastic *", "a photo of the cool *", "a close-up photo of a *", "a black and white photo of the *", "a painting of the *", "a painting of a *", "a pixelated photo of the *", "a sculpture of the *", "a bright photo of the *", "a cropped photo of a *", "a plastic *", "a photo of the dirty *", "a jpeg corrupted photo of a *", "a blurry photo of the *", "a photo of the *", "a good photo of the *", "a rendering of the *", "a * in a video game", "a photo of one *", "a doodle of a *", "a close-up photo of the *", "a photo of a *", "the origami *", "the * in a video game", "a sketch of a *", "a doodle of the *", "a origami *", "a low resolution photo of a *", "the toy *", "a rendition of the *", "a photo of the clean *", "a photo of a large *", "a rendition of a *", "a photo of a nice *", "a photo of a weird *", "a blurry photo of a *", "a cartoon *", "art of a *", "a sketch of the *", "a embroidered *", "a pixelated photo of a *", "itap of the *", "a jpeg corrupted photo of the *", "a good photo of a *", "a plushie *", "a photo of the nice *", "a photo of the small *", "a photo of the weird *", "the cartoon *", "art of the *", "a drawing of the *", "a photo of the large *", "a black and white photo of a *", "the plushie *", "a dark photo of a *", "itap of a *", "graffiti of the *", "a toy *", "itap of my *", "a photo of a cool *", "a photo of a small *", "a tattoo of the *.

\end{document}